\documentclass{article}

\PassOptionsToPackage{numbers, compress}{natbib}
\usepackage[preprint]{neurips_2026}

\usepackage[utf8]{inputenc} 
\usepackage[T1]{fontenc}    
\usepackage{hyperref}       
\usepackage{url}            
\usepackage{booktabs}       
\usepackage{amsfonts}       
\usepackage{nicefrac}       
\usepackage{microtype}      
\usepackage{xcolor}         
\usepackage{wrapfig}

\usepackage{amsmath}
\usepackage{amssymb}
\usepackage{mathtools}
\usepackage{amsthm}
\usepackage{algorithm}
\usepackage{algpseudocode}

\usepackage{makecell}
\usepackage{booktabs}

\usepackage[capitalize,noabbrev]{cleveref}
\usepackage{xcolor}
\usepackage[normalem]{ulem} 

\newcommand{\hiddencomment}[1]{}

\theoremstyle{plain}

\theoremstyle{definition}

\theoremstyle{remark}

\usepackage[textsize=tiny]{todonotes}

\title{Training Language Models via Neural Cellular Automata}

\author{%
  Dan Lee\thanks{Equal contribution.} \\
  UIUC \\
  \texttt{dhl9@illinois.edu} \\
  \And
  Seungwook Han\footnotemark[1] \\
  MIT, Improbable AI Lab \\
  \texttt{swhan@mit.edu} \\
  \And
  Akarsh Kumar \\
  MIT \\
  \And
  Pulkit Agrawal \\
  MIT, Improbable AI Lab \\
}

\begin{document}

\maketitle

\begin{abstract}
Pre-training is crucial for large language models (LLMs), as it is when most representations and capabilities are acquired.
However, natural language pre-training has problems: high-quality text is finite, contains human biases, and entangles knowledge with reasoning.
This raises a fundamental question: is natural language the only path to intelligence?
We propose using neural cellular automata (NCA) to generate synthetic, non-linguistic data for \textit{pre-pre-training} LLMs--training on synthetic-then-natural language.
NCA data exhibits rich spatiotemporal structure and statistics resembling natural language while being controllable and cheap to generate at scale.
We find that pre-pre-training on only 164M NCA tokens improves downstream language modeling by up to 6\% and accelerates convergence by up to 1.6$\times$ in 400M-3B models trained with Chinchilla optimal data budgets.
\textit{Surprisingly, this even outperforms pre-pre-training on 1.6B tokens of high-quality natural language data with more compute.} Investigating what drives transfer, we find that attention layers are the most transferable, and that optimal NCA complexity varies by domain: code benefits from simpler dynamics, while math and web text favor more complex ones. These results enable systematic tuning of the synthetic distribution to target domains. More broadly, our work opens a path toward more efficient models with fully synthetic pre-training.
\end{abstract}

\section{Introduction}
\label{sec:intro}
\begin{figure}[ht]
\centerline{\includegraphics[width=\columnwidth]{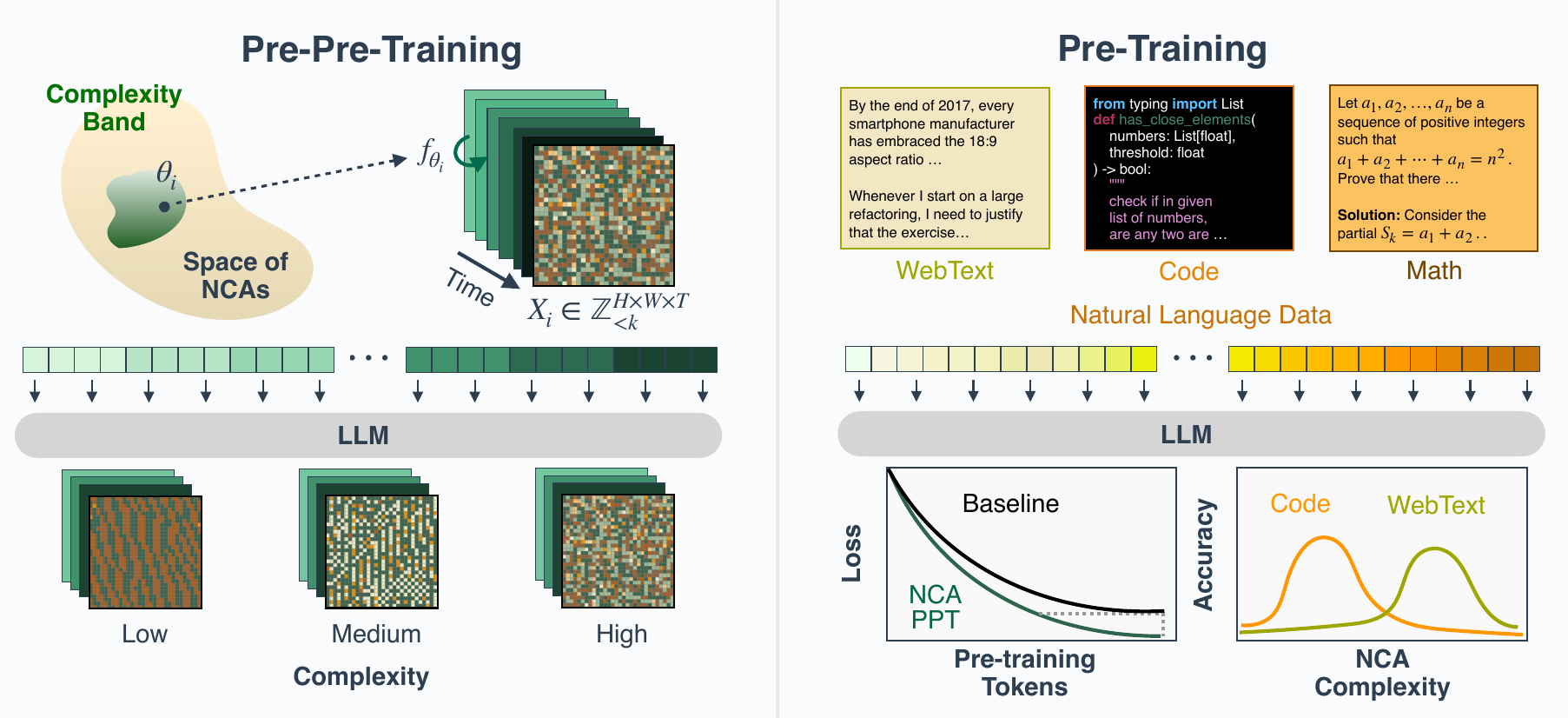}}
\caption{\textbf{Overview of NCA Pre-pre-training to Language Pre-training.}
We pre-pre-train a transformer with next-token prediction on the dynamics of neural cellular automata (NCA) sampled from selected complexity regions. We then conduct standard pre-training on natural language corpora. NCA pre-pre-training improves both validation perplexity and convergence speed on language pre-training. Interestingly, the optimal NCA distribution varies by downstream domain.}
\label{fig:teaser}
\end{figure}

Scale has transformed neural networks, enabling emergent abilities like reasoning~\citep{jaech2024openai,jiang2023latent, austin2021program} and in-context learning~\citep{brown2020language, wei2022emergentabilitieslargelanguage, zhao2024distributional} in large language models (LLMs). However, neural scaling laws predict that continued improvements require exponentially more data~\citep{kaplan2020scalinglawsneurallanguage}, which is nearing exhaustion by 2028 \citep{villalobos2022will}. Also, natural language inherits undesirable human biases and needs meticulous data curation before it is used for training~\citep{han2025generalintelligencerequiresrewardbased, an2024measuringgenderracialbiases}. This raises a fundamental question: \textbf{Is natural language the only path to learning useful representations?} In this paper, we explore an alternative path using synthetic data from cellular automata.

Our core hypothesis is that the efficacy of natural language for pre-training stems from its underlying structure, rather than its semantics. Text is a lossy record of human cognition and the world it describes, containing diverse kinds of structure, from reasoning traces to procedural instructions \citep{fractal-language, ruis2024procedural, cheng2025can, delétang2024languagemodelingcompression}. Next-token prediction on such data pressures models to internalize the latent computational processes for coherent continuations, fostering key capabilities of intelligence \citep{deletang2023language, jiang2023latent}.

If the key ingredient is exposure to various structures rather than language semantics, then richly structured non-linguistic data could also be effective for language pre-training. To investigate this hypothesis, we employ algorithmically generated data from neural cellular automata (NCA) \citep{mordvintsev2020growing} as a synthetic training substrate. NCA generalize systems like Conway's Game of Life \citep{gardner1970mathematical} by replacing fixed dynamics rules with neural networks and can generate diverse data distributions with simple local rules. This produces long-range spatio-temporal patterns (see Figure \ref{fig:teaser}) of arbitrary sizes that exhibit heavy-tailed, Zipfian token distributions (see Figure~\ref{fig:token-distributions} in Appendix \ref{app:data_distribution_analysis}) reminiscent of natural data. To systematically control this distribution, we propose a simple method to compute and sample the desired complexity of NCA tuned for the target domain.

We adopt a pre-pre-training framework\citep{hu2025circuitschomskyprepretrainingformal, jiang2023latent, bloem2025universalpretrainingiteratedrandom}: an initial phase of training on NCA dynamics that precedes standard pre-training on natural language.
Our ultimate vision is to pre-train entirely on clean synthetic data, followed by fine-tuning on a limited and curated corpora of natural language to acquire semantics \citep{han2025generalintelligencerequiresrewardbased}. The pre-pre-training framework serves as an early prototype of this paradigm, allowing us to measure how computational primitives learned from synthetic NCA transfer to language tasks. Our contributions are as follows: 

\begin{enumerate}
    \item \textbf{A fully synthetic pre-pre-training substrate that transfers to language modeling.}
    We propose neural cellular automata (NCA) as a fully algorithmic, non-linguistic data source for pre-pre-training. NCA pre-pre-training improves language modeling perplexity by up to 6\% and convergence speed by up to $1.6\times$ across web text, math, and code with 400M-3B models trained up to Chinchilla optimal token budgets. These performance gains translate to modest but statistically significant improvements on reasoning benchmarks, such as GSM8K and BigBench-Lite. Most surprisingly, NCA outperforms both web text (C4) and curated natural language data (FineWeb-Edu) as a pre-pre-training source.
    \item{\textbf{Synthetic pre-training enables domain-targeted data design.}}
We find that the optimal NCA complexity regime varies by downstream task: code benefits from lower-complexity rules while math and web text benefit from higher-complexity ones. NCAs' parametric structure offers a new lever to tune the complexity of training distributions to match the computational character of target domains.
    \item \textbf{Attention captures the most transferable priors.}
 The attention layers capture the most useful computational primitives, accounting for the majority of the transfer gains. Attention appears to be the carrier of transferable capabilities such as long-range dependency tracking and in-context learning, whereas MLPs may encode more domain-specific knowledge--making MLP transfer conditional on alignment between the synthetic and target domains. We further illustrate that models evidently learn different attention behaviors with and without NCA pre-pre-training.
 

\end{enumerate}

\section{Related Works}

Synthetic data include outputs generated by LLMs \citep{Nad__2025,wang2023selfinstructaligninglanguagemodels, xu2025wizardlmempoweringlargepretrained, mukherjee2023orcaprogressivelearningcomplex, li2023textbooksneediiphi15, lu2024mathgeniegeneratingsyntheticdata, wei2019edaeasydataaugmentation} and by simple algorithms. In this work, we pursue the latter, the non-linguistic approach.

\paragraph{Algorithm-Based Synthetic Data}
In vision, prior work used simple algorithmic procedures (e.g., OpenGL shader images, wavelet models, fractals) to generate synthetic training data~\citep{baradad2021learning, baradad2022procedural,kataoka2020pre}.
Despite lacking semantic content, these models learn representations that transfer well to real images.
\citet{baradad2021learning} argue that what matters is reproducing the statistical structure of the natural world, such as the approximate scale-invariance ~\citep{field1987relations} or the Zipfian distribution~\citep{zipf1949human,chan2022datadistributionalpropertiesdrive}. In the language domain, using algorithmically generated data is less common, but have been explored~\citep{saxton2019analysingmathematicalreasoningabilities,desai2015programsynthesisusingnatural,papadimitriou2023injecting}.
\citet{chiang2020pretraininglanguagemodelhuman, hu2025between} showed that pre-training LLMs on synthetic data generated by context-free grammars can be useful for language modeling. \citet{bloem2025universalpretrainingiteratedrandom} shows empirically that pre-training models on data generated by randomly initialized LSTMs can improve convergence speed for fine-tuning. Theoretically, they frame pre-training as an approximation of Solomonoff induction and demonstrate that data passed through computational processes can create depth that primes models for ``universal'' pattern recognition. Similarly, \citet{shinnick2025learnimagesproceduralwarmup} showed that warming up language models by training on abstract algorithmic sequences and formal languages like Dyck languages instill non-trivial structure in MLP and attention layers, accelerating the acquisition of  skills like long-context recall. \citet{berkovich2025lifegpt} and \citet{berkovich2025automatagpt} trained LLMs on 1D-elementary cellular automata, but did not study the usefulness of the learned representations for language. The key shortcoming of prior approaches with formal languages and random strings is that they are neither diverse nor controllable, offering no principled way to expand the space of tasks or to dial complexity up and down. NCAs address this gap. The parametric structure of NCA yields diverse dynamics and allows systematic control over complexity. This enables us to ask not only whether synthetic data can transfer, but what structural properties make it effective.

\paragraph{Shared Underlying Computation}
A growing body of work suggests that neural networks learn general computations that transfer across domains, raising the possibility that synthetic algorithmic data could instill such primitives directly. \citet{lu2022frozen} show that LLMs trained on natural language can transfer to seemingly unrelated domains like vision and protein folding.  \citet{huh2024platonicrepresentationhypothesis} illustrate that foundation models across different modalities are converging in representation, hinting that they learn a common structure.
Going further, \citet{mirchandani2023large} show that LLMs already have in-context learning capabilities for symbolic reasoning, numeric pattern continuation, and robotic control. These works cast LLMs as universal computation engines \citep{lu2022frozen, reid2022can, zhang2024intelligence}.

\paragraph{Emergent Complexity}
A central puzzle for algorithmic synthetic data is how simple procedures can give rise to data with rich structure, resembling the complexity of the real world. Researchers have developed various measures to quantify such complexity \citep{lloyd2001measures, mitchell2009complexity}, including Kolmogorov complexity \citep{kolmogorov1963tables}, sophistication \citep{mota2013sophistication}, and assembly index \citep{sharma2023assembly}. More recently, epiplexity was introduced as a complexity metric for computationally bounded observers \citep{finzi2026entropy}, demonstrating that simple deterministic rules can produce data useful for learning good representations. These works suggest that NCAs, despite having simple local rules, can generate arbitrarily complex structures when rolled out over long time horizons, making them a promising source of synthetic training data.

\section{Method}
\label{sec:method}

We study whether neural cellular automata (NCA) can create training data that teaches transferable computational priors useful for language modeling.

\subsection{Neural Cellular Automata (NCA)}
\label{subsec:nca}

NCAs generalize classical cellular automata such as Conway's Game of Life~\citep{WOLFRAM19841}. In the discrete setting used here, a grid evolves by synchronously updating each cell from its local neighborhood at the previous timestep. A small neural network parameterizes the shared update rule, allowing diverse rules to be sampled rather than hand-designed. Iterating a rule can produce structured spatiotemporal patterns (Figure ~\ref{fig:teaser} bottom left).

\paragraph{Random Discrete NCA.}
We use 2D discrete NCA on a $12{\times}12$ grid $g$ with periodic boundaries and state alphabet $n=10$, where each cell is represented as a 10-dimensional one-hot vector. The transition dynamics are governed by a neural network $f_\theta$ that maps each cell's $3{\times}3$ neighborhood to logits over next-cell states: $c^{(t+1)}_{i} \sim \mathrm{softmax}\!\left(f_\theta\!\left(N(c_i^{(t)})\right) / \tau\right)$,
where $c_i^{(t)}$ is the state of cell $i$ at time $t$, $N(c)$ denotes its neighborhood of the given cell, and $\tau=1\times10^{-3}$ is the sampling temperature to introduce small stochasticity. We parameterize $f_\theta$ as a $3{\times}3$ convolution (4 channels) followed by a cell-wise MLP with hidden size 16 and ReLU activation.

\paragraph{Complexity-based sampling.}

We outline our algorithm to generate diverse training data in \cref{alg:nca_gen} of Appendix \ref{appx:psuedocode}. For each sequence, we randomly initialize the parameters $\theta$ of the transition network and sample the initial grid $g^{(0)} \stackrel{\text{iid}}{\sim} \text{Unif}\{0,\dots,9\}^{12\times 12}$. These transition networks $f_{\theta}$ are not learned; they are frozen and unique dynamics functions for each rollout. This procedure yields a distribution over NCA dynamics ranging from trivially predictable states to highly chaotic ones.

To filter trajectories by structural complexity, we serialize each sequence as bytes and compute its gzip compressed-size ratio, $r = \texttt{compressed bytes} / \texttt{raw bytes} * 100$. We sample NCAs with trajectories of $r > 50\%$. gzip under the hood combines Lempel-Ziv compression ~\citep{lempelziv} with Huffman coding. Since Lempel-Ziv compression provides a computable upper bound on Kolmogorov complexity~\citep{LiVitanyi2019KC4}, gzip serves as a practical measure of sequence complexity: lower values indicate simpler, more predictable structure, whereas higher values indicate less compressible, more chaotic structure (see Figure~\ref{fig:teaser}). 

\paragraph{Tokenization.}
Given an NCA dynamics function $f_\theta$ and initial grid $g^{(0)}$, we generate a trajectory $(g^{(0)}, g^{(1)}, \dots, g^{(T)})$, then tokenize it patch-wise as in vision transformers~\citep{dosovitskiy2021imageworth16x16words}. We partition each grid $g^{(t)}$ into non-overlapping 2$\times$2 patches: $x^{(t)}_0, \dots, x^{(t)}_{P-1}$. With 10 possible states per cell and 4 cells per patch, there are $10^4$ distinct patches, which we map bijectively to a vocabulary of $10{,}000$ tokens (excluding special tokens). We serialize the $P=36$ patches of each grid in row-major order, wrap them with \texttt{<grid>} and \texttt{</grid>} delimiters, and concatenate across timesteps to form the final sequence $\mathbf{x} = (\texttt{<grid>}, x^{(0)}_0, \dots, x^{(0)}_{P-1}, \texttt{</grid>}, \texttt{<grid>}, x^{(1)}_0, \dots)$ capped at 2048 tokens.

\subsection{Training Objective and Interpretation}
We train a transformer autoregressively on the tokenized trajectory $\mathbf{x}=(x_1,\dots,x_N)$ using cross-entropy loss: $
\mathcal{L} = -\sum_{i=1}^{N} \log p_\phi(x_i \mid x_{<i})$. 
Because each sequence is generated by a separately sampled \(\theta\), next-token prediction encourages the model to infer and apply the trajectory's latent transition rule from context. This setup aligns with Bayesian accounts of in-context learning \citep{han2025emergenceeffectivenesstaskvectors, xie2022explanationincontextlearningimplicit}.

\section{Experimental Setup}
\label{sec:exp_setup}

\subsection{Training Paradigm}
We adopt a three-stage training paradigm \citep{shinnick2025transformers, hu2025circuitschomskyprepretrainingformal, bloem2025universalpretrainingiteratedrandom}:
\begin{enumerate}
    \item \textbf{Pre-pre-training:} An initial training phase designed to instill transferable computational priors before the main pre-training stage. In this work, we propose using synthetic, non-linguistic data (NCA trajectories) for pre-pre-training.
    \item \textbf{Pre-training:} Standard large-scale training on  natural language corpora (web text, code, or math) to acquire linguistic knowledge.
    \item \textbf{Fine-tuning:} Task-specific adaptation (e.g., instruction tuning).
\end{enumerate}
Our main focus is the transfer from stage 1 to stage 2---whether computational structure learned from synthetic data can accelerate and improve language model pre-training. As a secondary study, we examine stage 3, assessing how these gains carry over to downstream reasoning benchmarks.

\subsection{Setup}
We generate NCA data by randomly sampling neural network weights that define the transition rule. Each trajectory is thus produced by a unique rule, ensuring diversity across the training distribution. We conduct most of the main experiments with a 1.6B Llama-based transformer ~\citep{touvron2023llamaopenefficientfoundation} and scale up to 3B parameters.

We train on 164M NCA tokens sampled at the $50\%+$ gzip compressibility band, unless otherwise noted. We measure transfer by pre-training on four downstream corpora: OpenWebText~\citep{Gokaslan2019OpenWeb}, OpenWebMath~\citep{paster2023openwebmath}, CodeParrot~\citep{tunstall2022natural}, and FineWeb-Edu~\citep{fwe}, scaling up to the Chinchilla-optimal regime (20$\times$ the number of params) \citep{hoffmann2022trainingcomputeoptimallargelanguage}. We transfer all model weights except the embedding and output layers. All parameters are updated during pre-training. Additional details are in Appendix \ref{app:experimental-setup}.

For downstream reasoning benchmarks, we evaluate on GSM8K \citep{cobbe2021training}, HumanEval \citep{chen2021evaluating}, and BigBench-Lite \citep{srivastava2023beyond}. We fine-tune the models on the train set for instruction following for GSM8K and BigBench-Lite. We provide the fine-tuning details in Appendix~\ref{app:instruction-ft-setup}.

\subsection{Baselines}
We compare four baseline conditions: (i) \textbf{scratch}, in which the model is randomly initialized and trained directly on the language corpus; (ii) \textbf{Dyck pre-pre-training}, using the \(k\)-Dyck formal language studied by \citet{hu2025circuitschomskyprepretrainingformal}; (iii) \textbf{C4 pre-pre-training}; and (iv) \textbf{FineWeb-Edu pre-pre-training}. For Dyck, we use \(k=128\) and unbounded potential depth. Within each training stage, we match token counts, optimizer steps, and total FLOPs across conditions; only the pre-pre-training data source varies.

\subsection{Hyperparameters}
For both our pre-pre-training and baseline runs, we perform a grid search over learning rate and weight decay and use the best hyperparameters for each method. For pre-training, we train four separate models for a single epoch on each dataset: Fineweb-Edu (32B and 98B tokens), OpenWebText (9B tokens), OpenWebMath (4B tokens), and CodeParrot (13B tokens). We report the detailed hyperparameters in Table~\ref{tab:hyperparams} in Appendix~\ref{app:experimental-setup}.

For downstream reasoning benchmarks, we use models pre-trained on OpenWebText, OpenWebMath, and CodeParrot for Big-Bench-Lite, GSM8K, and HumanEval, respectively: matching the benchmark domain approximately to the pre-training domain. We report details on fine-tuning in Appendix~\ref{app:instruction-ft-setup}.

\subsection{Evaluation Metrics for Transfer}
We evaluate transfer from pre-pre-training to pre-training using validation perplexity on a held-out set and convergence speed~\citep{bloem2025universalpretrainingiteratedrandom, hu2025circuitschomskyprepretrainingformal, kaplan2020scalinglawsneurallanguage}. Convergence speed is measured as the ratio of tokens to reach a base model's final perplexity relative to the tokens required by the NCA pre-pre-trained model, including any pre-pre-training tokens. Adapting~\citet{hu2025circuitschomskyprepretrainingformal}, we compute this as,
$S^{m}_{\text{NCA}} = \frac{T_{\text{PPT}}^{m} + T_{\text{PT}}^{m}}{T_{\text{PPT}}^{\text{NCA}} + T_{\text{PT}}^{\text{NCA}}}$,
where $T_{\text{PPT}}^{m}$ and $T_{\text{PT}}^{m}$ are the pre-pre-training and pre-training tokens that model $m$ needs to reach the final loss. The baseline $m$ is typically the scratch model, for which $T_{\text{PPT}}^{\text{scratch}} = 0$.

On downstream reasoning tasks, our primary evaluation metric is pass accuracy with multiple decodings or pass@$k$ \citep{chen2021evaluatinglargelanguagemodels}. For BigBench tasks, we compute accuracy on a normalized basis to adjust for random guessing associated with multiple-choice questions \citep{srivastava2023beyond}.





\begin{figure*}[t]
\centering
\includegraphics[width=\textwidth]{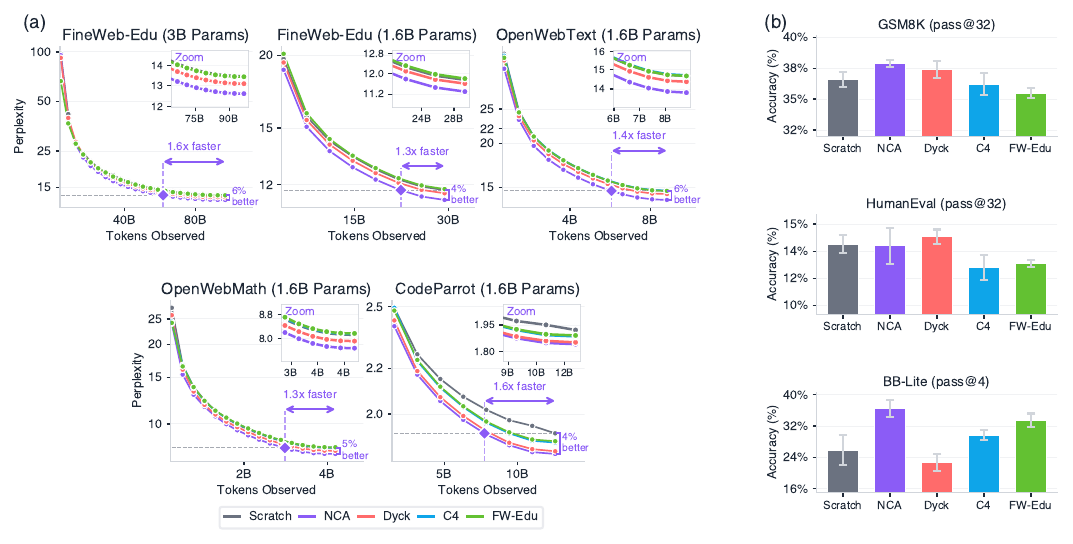}
\vskip -0.1 in
\caption{\textbf{(a) NCA pre-pre-training improves and accelerates LM pre-training across domains.} Validation perplexity for 3B models on FineWeb-Edu and 1.6B models on FinewebEdu, OpenWebText, OpenWebMath, and CodeParrot. NCA outperforms scratch, Dyck, and even C4 and FineWeb-Edu baselines, yielding 1.3–1.6x faster convergence and 4–6\% lower final perplexity. \textbf{(b) NCA pre-pre-training provides modest but statistically significant improvement on GSM8k and Bigbench.} Pass@k accuracy and standard deviation error bars for 1.6B language models on reasoning benchmarks.}
\label{fig:combined_perplexity}
\vskip -0.25 in
\end{figure*}

\section{Results}
\label{sec:results}

We present results on the impact of NCA pre-pre-training on downstream language modeling, and analyze how transfer varies across scale and data complexity.

\subsection{NCA Pre-pre-training Improves Language Modeling}
\label{subsec:nca_transfer}

We compare the language modeling performance against three baselines: (i) no pre-pre-training (``scratch''), (ii) pre-pre-training on another synthetic language Dyck \citep{hu2025circuitschomskyprepretrainingformal}, and (iii) pre-pre-training on natural language datasets C4 and FineWeb-Edu. 

\begin{figure*}[t]
\centering
\includegraphics[width=\textwidth]{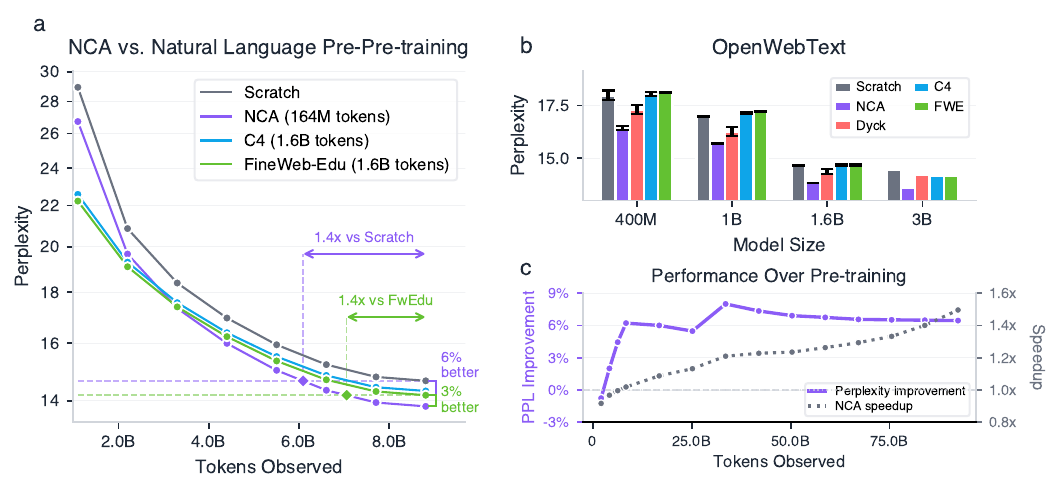}
\caption{\textbf{(a) 164M tokens of NCA pre-pre-training improves convergence speed and perplexity versus 1.6B of natural language pre-pre-training.} Comparison of convergence and final validation perplexity on OpenWebText of models pre-pretrained on 164M tokens of NCA with reinitialized embeddings versus models pre-pretrained on 1.6B tokens on C4 and FineWeb-Edu with embeddings transferred. \textbf{(b) NCA pre-pre-training improvements persist across model scale.} Final validation perplexity and standard deviations (seeds) across all baselines and model scales from 400M to 3B. \textbf{(c) Perplexity gains remain stable and convergence speedup grows throughout pre-training. }We compute the perplexity improvement and convergence speedup for NCA pre-pretrained models versus scratch models (3B) during 98B tokens of Fineweb-Edu pre-training.}
\label{fig:scaling}
\end{figure*}

As seen in Figure~\ref{fig:scaling}b, across model scales (400M, 600M, 1.6B, 3B) and multiple random seeds, NCA pre-pre-trained models consistently outperform all four baselines (Scratch, Dyck, C4 and FineWeb-Edu). On OpenWebText, the best-performing NCA pre-pre-trained 400M model improves downstream perplexity upon the scratch baseline by \textbf{8.6\%}, and the 3B model improves by  \textbf{5.8\%}. On FineWeb-Edu, the gains grow with scale: moving from 1.6B to 3B (with data budget scaled accordingly), perplexity improvement increases from 4\% to 6\% and convergence speedup from 1.3x to 1.6x. Figure~\ref{fig:scaling}c shows that for the 3B model trained on 98B tokens, the NCA model's convergence advantage over the scratch baseline grows with the token budget, and the perplexity improvement is maintained throughout training. This suggests that NCA's benefits are not fragile artifacts of initialization but persist across model and data scale.

The most surprising observation is that \textbf{\textit{pre-pre-training on NCA outperforms pre-pre-training on natural language (C4 and FineWeb-Edu) under matched token and compute budgets}} (Figure~\ref{fig:combined_perplexity}). We strengthen this comparison by giving the FineWeb-Edu baseline substantially more data---1.6B versus 164M tokens---and retaining its trained embeddings during transfer (Figure~\ref{fig:scaling}a). Even with these advantages, NCA models improve upon the FineWeb-Edu baseline by 5\% on perplexity and 1.4$\times$ on training convergence.

Transfer also extends to math and code (Figure~\ref{fig:combined_perplexity}a): NCA pre-pre-training reduces validation perplexity on OpenWebMath and CodeParrot by \(4\)--\(5\%\) and yields up to a \(1.6\times\) convergence speedup. The advantage persists and grows during training, indicating greater token efficiency rather than a transient initialization effect.

\subsection{Impact of NCA Pre-Pre-Training on Downstream Language and Reasoning Tasks}
\label{subsec:downstream_tasks}
While perplexity and convergence speed are robust metrics for language modeling, we further evaluate whether the improvements from the NCA pre-pre-training manifest in downstream reasoning. We present the results on GSM8K, HumanEval, and BigBench-Lite in \cref{fig:combined_perplexity}b. Given the model scale (1.6B) and limited downstream data budget, absolute performance remains limited; however, we observe \textbf{statistically significant improvements ($p<0.05$)} on GSM8K and Bigbench-Lite compared to the scratch and natural language baselines. 

On GSM8K, NCA pre-pre-training improves accuracy from 36.6\% to 37.9\% at pass@32. On HumanEval, NCA pre-pre-training improves pass@1 but the advantage is not statistically significant. Interestingly, Dyck pre-pre-training is competitive with NCA on HumanEval at higher pass@$k$. This likely reflects the structural overlap between Dyck languages 
and code, both of which require tracking nesting logic and delimiter patterns. On BigBench-Lite, NCA pre-pre-training outperforms markedly at higher $k$, reaching 36.5\% at pass@4 compared to 29.7\% for C4 and 25.9\% for the scratch baseline. In Appendix~\ref{app:coding_benchmarks}, we provide additional experiments for evaluations on the Mostly Basic Python Programs (MBPP) programming benchmark \citep{mbpp2021}, where we observe that NCA models improve upon the scratch baseline by $4.55\%$ at $\textit{pass@32}$. Together, these results suggest that NCA pre-pre-training can offer modest gains in downstream reasoning, but we leave a more thorough characterization of when and why these gains emerge to future work. A detailed comparison over different pass@$k$ (\cref{table:combined-results}) and full t-tests and p-values (\cref{table:statistical-significance}) are reported in Appendix~\ref{app:statistical_significance}.

\begin{figure}[t]
\centerline{\includegraphics[width=\columnwidth]{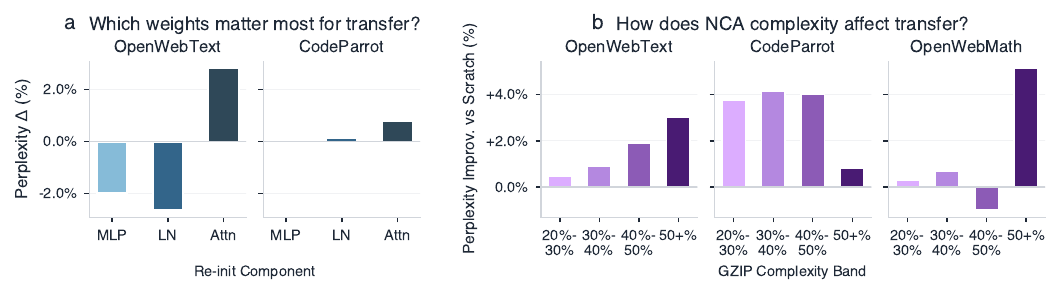}}
\caption{Across ablations, we report the validation perplexity change (more negative is better) on Openwebtext relative to (a) NCA pre-pretrained models with no reinitializations and (b) scratch model. \textbf{(a) Attention weights are most crucial for positive transfer.} Re-initializing attention causes the largest degradation across both OpenWebText and CodeParrot, while MLP and LayerNorm effects are domain-dependent. \textbf{(b) Optimal NCA complexity is domain-dependent.} Text benefits from higher-complexity data (50\%+), while Code peaks at intermediate complexity (30–40\%). This suggests that matching synthetic data complexity to the target domain is needed to maximize transfer. 
}
\label{fig:ablations_attn_gzip}
\end{figure}

\subsection{What Drives Transfer?}
\label{subsec:what_drives}

The preceding results establish that NCA pre-pre-training improves both language modeling and reasoning. We now investigate the mechanism: which model components capture the transferable structure, and what properties of the synthetic data control transfer effectiveness?

\subsubsection{Attention captures the most transferable primitives.}
\label{subsubsec:attn_transfer}

To isolate which model components carry the transfer signal, we selectively re-initialize subsets of weights after NCA pre-pre-training and measure the impact on language modeling. As shown in Figure~\ref{fig:ablations_attn_gzip}a, re-initializing attention weights causes the largest degradation in transfer across all configurations, indicating that attention captures the most transferable computational primitives.

Other components are more domain-dependent. On OpenWebText, retaining MLP and LayerNorm weights hurts transfer, suggesting that they encode NCA-specific statistics that interfere with language learning. On CodeParrot, retaining these weights has negligible effect. Thus, attention transfers across the tested domains, whereas MLP and LayerNorm transfer depends on source-to-target alignment.

\begin{wrapfigure}[20]{r}{0.5\columnwidth}
\centering
\includegraphics[width=0.48\columnwidth]{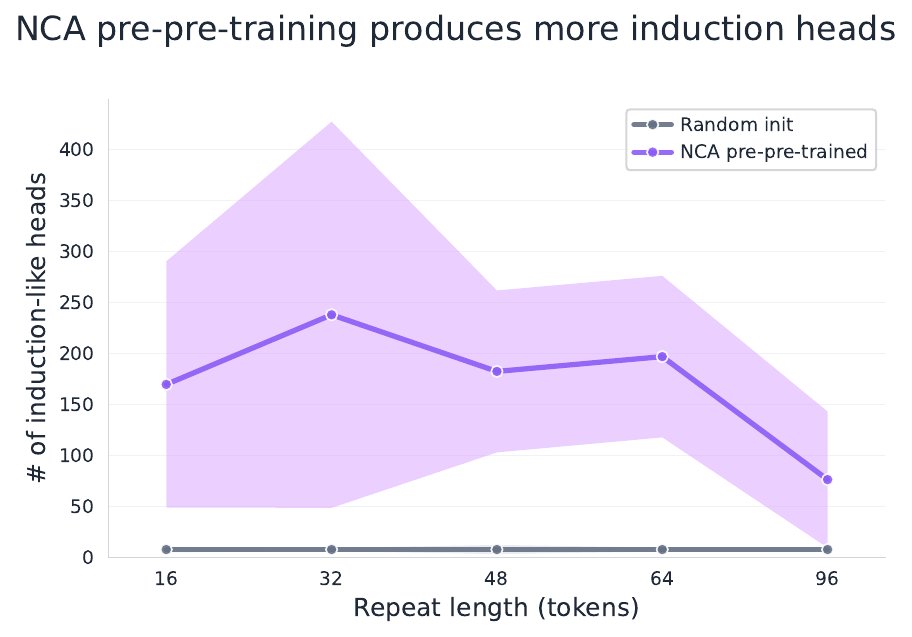}
\caption{Across repeat lengths, \textbf{NCA pre-pre-trained models contain substantially more induction-like heads} than randomly initialized models. We plot the means and $\pm 1$ standard deviation across four model seeds.}
\label{fig:induction_heads}
\end{wrapfigure}

This interpretation agrees with concurrent work locating transferable structure primarily in attention during synthetic pre-training~\citep{shinnick2025transformers,shinnick2025learnimagesproceduralwarmup}. It is also consistent with analyses of Mixture-of-Experts architectures suggesting that MLP reinforces knowledge than reasoning~\citep{jelassi2025mixtureparrotsexpertsimprove}. Together, these findings suggest a functional division: attention learns general mechanisms for tracking dependencies and inferring latent rules, whereas MLPs store more domain-specific patterns and statistics.

To test whether NCA pre-pre-training strengthens a specific in-context mechanism, we probe the prefix-matching component of induction heads~\citep{olsson2022context}. NCA-pre-trained models contain $22\times$ as many above-null prefix-matching heads as randomly initialized models (Figure~\ref{fig:induction_heads}). This provides mechanistic evidence that NCA pre-pre-training places a canonical in-context computation primitive in the weights rather than merely providing a favorable optimization warmup. Appendix~\ref{app:induction_heads} details the probe. To test whether the broader transfer gains can be explained solely by improved initialization statistics, we also compare against first-order-statistics-based mimetic initialization. NCA pre-pre-training outperforms this control (Appendix~\ref{app:mimetic}), indicating that its benefits extend beyond initialization alone.

\subsubsection{Data complexity modulates transfer and the optimum is domain-dependent.}

Having established that attention carries the most transferable signal, we next ask: what properties of NCA data affect the transfer? We analyze complexity along two complementary axes: gzip compressibility (as an upper bound to Kolmogorov complexity) and alphabet size $n$ (size of the state space).

\paragraph{Complexity via gzip.} We generate NCA trajectories of varying gzip compressibility bands (20–30\%, 30–40\%, 40–50\%, 50\%+). Smaller compression ratios imply regular, low-entropy dynamics that are more predictable, whereas larger compression ratios generate more diverse and unpredictable trajectories with richer spatiotemporal structure, as illustrated in \cref{fig:teaser}.

As shown in Figure~\ref{fig:ablations_attn_gzip}b, the optimal complexity band varies by downstream domain. OpenWebText benefits from less compressible (more complex) NCA rules in the 50\%+ band, while CodeParrot shows a sweet spot at intermediate complexity (30–40\% gzip). Strikingly, this aligns with the intrinsic complexity of the target corpora themselves in Figure~\ref{fig:token-distributions} of Appendix~\ref{app:data_distribution_analysis}: OpenWebText and OpenWebMath have gzip ratios of 60–70\%, whereas CodeParrot is substantially more compressible at 32\%. The correlation is somewhat direct. Domains with higher intrinsic complexity benefit from higher-complexity synthetic data, and vice versa. This suggests a plausible and practical principle: \textbf{matching the complexity of synthetic pre-training data to the target domain maximizes transfer}.

\begin{wrapfigure}{r}{0.5\columnwidth}
\begin{center}
    \includegraphics[width=0.48\columnwidth]{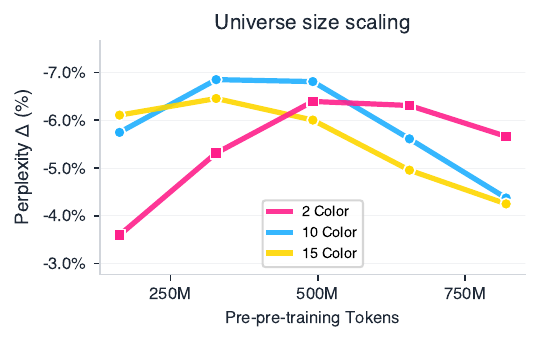}
\end{center}
\caption{\textbf{NCA alphabet size changes scaling behavior.} Smaller alphabets scale better.}
\label{fig:ablations_universe}
\end{wrapfigure}

\textbf{Rule space expressiveness via alphabet size.} We vary the NCA state alphabet $n \in \{2, 10, 15\}$, which controls the diversity of possible local interactions. As shown in Figure~\ref{fig:ablations_universe}, larger alphabets ($n=10, 15$) exhibit diminishing returns: performance improves the most at an intermediate NCA token budget, then the improvement gap narrows. Surprisingly, the smallest alphabet ($n=2$) scales most favorably, continuing to improve where larger alphabets plateau. 

As seen in Figure~\ref{fig:nca_gzip} in Appendix~\ref{app:data_distribution_analysis}, when increasing $n$, the resulting NCA data naturally becomes more complex. This result suggests that although larger rule spaces can express more complex dynamics, better guidance is necessary to sample a diverse set of NCAs that optimally transfer to language. Thus, constraining the space to $n{=}2$ may paradoxically help by concentrating samples on dynamics with more consistent and transferable structure.

These results reveal that it is not simply ``more NCA data is better.'' The complexity of synthetic data, both gzip and alphabet size, must be calibrated. This offers a lever unavailable in natural language pre-training: the ability to tune the training distribution to match the computational character of target domains.

We further discuss the implications for domain-targeted pre-training in Section~\ref{sec:discussion}.

\section{Discussion}
\label{sec:discussion}

\paragraph{Why should we expect transfer?}
NCA data are substantially different from natural language and are generated by deterministic processes, prompting the question of why one should expect transfer at all? Yet our results from Section \ref{sec:results} and \cref{fig:loss_during_train} in Appendix \ref{app:owt_during_nca} shows transfer occurs in both directions during training—NCA training improves language loss and vice versa, suggesting a shared underlying structure.

NCAs can provide a direct signal for in-context rule inference.. In natural language, models can exploit semantic shortcuts or co-occurrence priors~\citep{abbas2023semdedupdataefficientlearningwebscale,geirhos2020shortcut}. In contrast, each NCA sequence instead follows a hidden transition rule that must be inferred from context~\citep{kirsch2022general}, without any semantic knowledge to fall back on. Such rule inference is also central to modeling language~\citep{brown2020language, wei2022emergentabilitieslargelanguage, dong2024survey}, math, and code \citep{xie2022explanationincontextlearningimplicit, garg2023transformerslearnincontextcase, cook2025programming}. Because NCAs encompass a broad, universal class of computable functions \citep{Copeland2012}, some of which realize Turing-complete systems \citep{Rendell2002, wolfram2003new}, their diversity makes memorization impractical and favors general rule-inference mechanisms \citep{li2024languagemodelslearncontext}. This framing is supported by our findings in Section~\ref{subsubsec:attn_transfer} that attention carries the most transferable structure. Training on NCA could help with the formation of induction heads (copy-and-apply behavior), which \citet{olsson2022context} identify as the important substrate for in-context learning. A complementary motivation comes from \textit{epiplexity} \citep{finzi2026entropy}: although classical information theory suggests deterministic transformations cannot increase information content \citep{Polyanskiy_Wu_2025}, this assumes a computationally unbounded observer. For computationally bounded observers, \citet{finzi2026entropy} show that simple local rules give rise to emergent structures (e.g., gliders, collisions) that cannot be brute-force simulated within a finite model's capacity. This forces the model to learn coarser-grained abstractions, mirroring the representational pressure that drives language modeling.

\paragraph{Why are 164M tokens of automata better than 1.6B tokens of text?}
We propose two explanations for this data-efficiency gap. First, early in training, language models learn shallow, local patterns before more complex structures~\citep{evanson2023language,chen2023sudden}. C4 and FineWeb-Edu may therefore devote more capacity to surface regularities than to broadly transferable long-range dependencies and in-context learning. Each NCA sequence, by contrast, requires the model to infer and consistently apply a latent rule. Second, NCA contributes diversity orthogonal to more text: natural language contains substantial redundancy in patterns and topics~\citep{abbas2023semdedupdataefficientlearningwebscale}, whereas each NCA sequence represents a distinct function, increasing diversity per token.

\paragraph{Beyond one-size-fits-all pre-training}

Our complexity ablations reveal a nuanced picture that the optimal distribution for training varies by downstream domain. In \cref{fig:ablations_attn_gzip}b, we observed that code benefits from lower-complexity NCA rules, while web text and math benefit from higher-complexity ones, suggesting these domains encode computations of measurably different character. This opens a new axis of control. Rather than treating training data as fixed, we can tune the structures of synthetic data to match the target domain. Unlike grammar-based synthetic tasks, where each formal grammar defines a task with fixed structural complexity, NCAs provide a continuous and tunable spectrum of complexity within a single generator family. If researchers can craft distributions that embody the primitives a domain requires (e.g., rigid state-tracking for code \citep{li2025codeiocondensingreasoningpatterns}, richer long-range dependencies for genomic sequences \citep{wu2025generatorlongcontextgenerativegenomic}), they can instill these capabilities directly, without scaling to trillions of general-purpose tokens.
The result could aid the development of specialized, small language models \citep{belcak2025smalllanguagemodelsfuture} that are more efficient to train and deploy---trained not on more data, but on better-matched data.

\paragraph{Limitations and open problems}
\label{subsec:limits_scaling_nca}
We have demonstrated the effectiveness of NCA pre-pre-training up to 3B parameters and slightly beyond Chinchilla-optimal token budgets (98B tokens), and find that the gains in convergence speed and language modeling persist with model and data scale. Extending these experiments to the multi-trillion token regime of frontier LMs would be a valuable direction for future work. A related question is whether NCA data can serve not just as a pre-pre-training signal but as a scalable substitute for natural language pre-training. For larger alphabet sizes ($n=10,15$), we observe a reverse U-shaped trend: downstream improvement peaks at an intermediate token budget and plateaus beyond it—reinforcing that effective synthetic pre-training depends critically on structural choices in the generator, not merely on scale.
This points to a key open problem: developing principled methods to guide synthetic generators toward structures that match target domains. Our complexity results show that such matching matters, but gzip compressibility and alphabet size are only two lenses on a multifaceted property. Characterizing which axes of complexity matter for which domains could unlock fully synthetic pre-training at scale. NCA is one point in the vast space of possible synthetic generators; the key insight is not that NCA specifically is optimal, but that structured synthetic data with appropriate complexity can provide meaningful pre-training signal without any linguistic content. The question is no longer whether synthetic pre-training can work, but how to design distributions that maximize what models learn.

\section*{Author Contributions}
\textbf{Dan Lee} co-lead the project and contributed to all aspects of experiments and writing.

\textbf{Seungwook Han} co-lead the project and contributed to all aspects of experiments and writing.

\textbf{Akarsh Kumar} supported this project, contributed to the design of the experiments and significantly to the writing.

\textbf{Pulkit Agrawal} advised the development of the project idea from inception and contributed significantly to the writing.

\section*{Acknowledgment}
We want to express our gratitude to Zachary Shinnick, Phillip Isola, Yoon Kim, Ryan Bahlous-Boldi, Idan Shenfeld, Nitish Dashora, Jyo Pari, Aldo Pareja, and members of the Improbable
AI lab for the helpful discussion on the paper. The
research was supported in part by NSF CSGrad4US Fellowship and Google. The research was sponsored by the Army Research Office and was accomplished under Grant Number W911NF-23-1-0277 and W911NF-21-1-0328. Also, the research was sponsored by the Department of the Navy, Office of Naval Research under Grant Number N00014-22-1-2740. The views and conclusions contained in this document are those of the authors and should not
be interpreted as representing the official policies, either
expressed or implied, of the Army Research Office, Naval
Research Office, or the U.S. Government.

\newpage

\bibliography{references}

\begin{thebibliography}{88}
\providecommand{\natexlab}[1]{#1}
\providecommand{\url}[1]{\texttt{#1}}
\expandafter\ifx\csname urlstyle\endcsname\relax
  \providecommand{\doi}[1]{doi: #1}\else
  \providecommand{\doi}{doi: \begingroup \urlstyle{rm}\Url}\fi

\bibitem[Abbas et~al.(2023)Abbas, Tirumala, Simig, Ganguli, and Morcos]{abbas2023semdedupdataefficientlearningwebscale}
Abbas, A., Tirumala, K., Simig, D., Ganguli, S., and Morcos, A.~S.
\newblock Semdedup: Data-efficient learning at web-scale through semantic deduplication, 2023.
\newblock URL \url{https://arxiv.org/abs/2303.09540}.

\bibitem[An et~al.(2024)An, Huang, Lin, and Tai]{an2024measuringgenderracialbiases}
An, J., Huang, D., Lin, C., and Tai, M.
\newblock Measuring gender and racial biases in large language models, 2024.
\newblock URL \url{https://arxiv.org/abs/2403.15281}.

\bibitem[Austin et~al.(2021{\natexlab{a}})Austin, Odena, Nye, Bosma, Michalewski, Dohan, Jiang, Cai, Terry, Le, et~al.]{austin2021program}
Austin, J., Odena, A., Nye, M., Bosma, M., Michalewski, H., Dohan, D., Jiang, E., Cai, C., Terry, M., Le, Q., et~al.
\newblock Program synthesis with large language models.
\newblock \emph{arXiv preprint arXiv:2108.07732}, 2021{\natexlab{a}}.

\bibitem[Austin et~al.(2021{\natexlab{b}})Austin, Odena, Nye, Bosma, Michalewski, Dohan, Jiang, Cai, Terry, Le, and Sutton]{mbpp2021}
Austin, J., Odena, A., Nye, M.~I., Bosma, M., Michalewski, H., Dohan, D., Jiang, E., Cai, C.~J., Terry, M., Le, Q.~V., and Sutton, C.
\newblock Program synthesis with large language models.
\newblock \emph{CoRR}, abs/2108.07732, 2021{\natexlab{b}}.
\newblock URL \url{https://arxiv.org/abs/2108.07732}.

\bibitem[Baradad et~al.(2022)Baradad, Chen, Wulff, Wang, Feris, Torralba, and Isola]{baradad2022procedural}
Baradad, M., Chen, R., Wulff, J., Wang, T., Feris, R., Torralba, A., and Isola, P.
\newblock Procedural image programs for representation learning.
\newblock \emph{Advances in Neural Information Processing Systems}, 35:\penalty0 6450--6462, 2022.

\bibitem[Baradad~Jurjo et~al.(2021)Baradad~Jurjo, Wulff, Wang, Isola, and Torralba]{baradad2021learning}
Baradad~Jurjo, M., Wulff, J., Wang, T., Isola, P., and Torralba, A.
\newblock Learning to see by looking at noise.
\newblock \emph{Advances in Neural Information Processing Systems}, 34:\penalty0 2556--2569, 2021.

\bibitem[Belcak et~al.(2025)Belcak, Heinrich, Diao, Fu, Dong, Muralidharan, Lin, and Molchanov]{belcak2025smalllanguagemodelsfuture}
Belcak, P., Heinrich, G., Diao, S., Fu, Y., Dong, X., Muralidharan, S., Lin, Y.~C., and Molchanov, P.
\newblock Small language models are the future of agentic ai, 2025.
\newblock URL \url{https://arxiv.org/abs/2506.02153}.

\bibitem[Berkovich \& Buehler(2025)Berkovich and Buehler]{berkovich2025lifegpt}
Berkovich, J.~A. and Buehler, M.~J.
\newblock Lifegpt: Topology-agnostic generative pretrained transformer model for cellular automata.
\newblock \emph{npj Artificial Intelligence}, 1\penalty0 (1):\penalty0 23, 2025.

\bibitem[Berkovich et~al.(2025)Berkovich, David, and Buehler]{berkovich2025automatagpt}
Berkovich, J.~A., David, N.~S., and Buehler, M.~J.
\newblock Automatagpt: Forecasting and ruleset inference for two-dimensional cellular automata.
\newblock \emph{arXiv preprint arXiv:2506.17333}, 2025.

\bibitem[Bloem(2025)]{bloem2025universalpretrainingiteratedrandom}
Bloem, P.
\newblock Universal pre-training by iterated random computation, 2025.
\newblock URL \url{https://arxiv.org/abs/2506.20057}.

\bibitem[Brown et~al.(2020)Brown, Mann, Ryder, Subbiah, Kaplan, Dhariwal, Neelakantan, Shyam, Sastry, Askell, Agarwal, Herbert-Voss, Krueger, Henighan, Child, Ramesh, Ziegler, Wu, Winter, Hesse, Chen, Sigler, Litwin, Gray, Chess, Clark, Berner, McCandlish, Radford, Sutskever, and Amodei]{brown2020language}
Brown, T.~B., Mann, B., Ryder, N., Subbiah, M., Kaplan, J., Dhariwal, P., Neelakantan, A., Shyam, P., Sastry, G., Askell, A., Agarwal, S., Herbert-Voss, A., Krueger, G., Henighan, T., Child, R., Ramesh, A., Ziegler, D.~M., Wu, J., Winter, C., Hesse, C., Chen, M., Sigler, E., Litwin, M., Gray, S., Chess, B., Clark, J., Berner, C., McCandlish, S., Radford, A., Sutskever, I., and Amodei, D.
\newblock Language models are few-shot learners, 2020.
\newblock URL \url{https://arxiv.org/abs/2005.14165}.

\bibitem[Chan et~al.(2022)Chan, Santoro, Lampinen, Wang, Singh, Richemond, McClelland, and Hill]{chan2022datadistributionalpropertiesdrive}
Chan, S. C.~Y., Santoro, A., Lampinen, A.~K., Wang, J.~X., Singh, A., Richemond, P.~H., McClelland, J., and Hill, F.
\newblock Data distributional properties drive emergent in-context learning in transformers, 2022.
\newblock URL \url{https://arxiv.org/abs/2205.05055}.

\bibitem[Chen et~al.(2023)Chen, Shwartz-Ziv, Cho, Leavitt, and Saphra]{chen2023sudden}
Chen, A., Shwartz-Ziv, R., Cho, K., Leavitt, M.~L., and Saphra, N.
\newblock Sudden drops in the loss: Syntax acquisition, phase transitions, and simplicity bias in mlms.
\newblock \emph{arXiv preprint arXiv:2309.07311}, 2023.

\bibitem[Chen(2021)]{chen2021evaluating}
Chen, M.
\newblock Evaluating large language models trained on code.
\newblock \emph{arXiv preprint arXiv:2107.03374}, 2021.

\bibitem[Chen et~al.(2021)Chen, Tworek, Jun, Yuan, de~Oliveira~Pinto, Kaplan, Edwards, Burda, Joseph, Brockman, Ray, Puri, Krueger, Petrov, Khlaaf, Sastry, Mishkin, Chan, Gray, Ryder, Pavlov, Power, Kaiser, Bavarian, Winter, Tillet, Such, Cummings, Plappert, Chantzis, Barnes, Herbert-Voss, Guss, Nichol, Paino, Tezak, Tang, Babuschkin, Balaji, Jain, Saunders, Hesse, Carr, Leike, Achiam, Misra, Morikawa, Radford, Knight, Brundage, Murati, Mayer, Welinder, McGrew, Amodei, McCandlish, Sutskever, and Zaremba]{chen2021evaluatinglargelanguagemodels}
Chen, M., Tworek, J., Jun, H., Yuan, Q., de~Oliveira~Pinto, H.~P., Kaplan, J., Edwards, H., Burda, Y., Joseph, N., Brockman, G., Ray, A., Puri, R., Krueger, G., Petrov, M., Khlaaf, H., Sastry, G., Mishkin, P., Chan, B., Gray, S., Ryder, N., Pavlov, M., Power, A., Kaiser, L., Bavarian, M., Winter, C., Tillet, P., Such, F.~P., Cummings, D., Plappert, M., Chantzis, F., Barnes, E., Herbert-Voss, A., Guss, W.~H., Nichol, A., Paino, A., Tezak, N., Tang, J., Babuschkin, I., Balaji, S., Jain, S., Saunders, W., Hesse, C., Carr, A.~N., Leike, J., Achiam, J., Misra, V., Morikawa, E., Radford, A., Knight, M., Brundage, M., Murati, M., Mayer, K., Welinder, P., McGrew, B., Amodei, D., McCandlish, S., Sutskever, I., and Zaremba, W.
\newblock Evaluating large language models trained on code, 2021.
\newblock URL \url{https://arxiv.org/abs/2107.03374}.

\bibitem[Cheng et~al.(2025)Cheng, Cao, Pishdad, Cao, and Cheung]{cheng2025can}
Cheng, Z., Cao, M., Pishdad, L., Cao, Y., and Cheung, J.~C.
\newblock Can llms reason abstractly over math word problems without cot? disentangling abstract formulation from arithmetic computation.
\newblock In \emph{Proceedings of the 2025 Conference on Empirical Methods in Natural Language Processing}, pp.\  14317--14344, 2025.

\bibitem[Chiang \& Lee(2020)Chiang and Lee]{chiang2020pretraininglanguagemodelhuman}
Chiang, C.-H. and Lee, H.
\newblock Pre-training a language model without human language, 2020.
\newblock URL \url{https://arxiv.org/abs/2012.11995}.

\bibitem[Cobbe et~al.(2021)Cobbe, Kosaraju, Bavarian, Chen, Jun, Kaiser, Plappert, Tworek, Hilton, Nakano, et~al.]{cobbe2021training}
Cobbe, K., Kosaraju, V., Bavarian, M., Chen, M., Jun, H., Kaiser, L., Plappert, M., Tworek, J., Hilton, J., Nakano, R., et~al.
\newblock Training verifiers to solve math word problems.
\newblock \emph{arXiv preprint arXiv:2110.14168}, 2021.

\bibitem[Cook et~al.(2025)Cook, Sapora, Ahmadian, Khan, Rocktaschel, Foerster, and Ruis]{cook2025programming}
Cook, J., Sapora, S., Ahmadian, A., Khan, A., Rocktaschel, T., Foerster, J., and Ruis, L.
\newblock Programming by backprop: Llms acquire reusable algorithmic abstractions during code training.
\newblock \emph{arXiv preprint arXiv:2506.18777}, 2025.

\bibitem[Copeland(2012)]{Copeland2012}
Copeland, B.~J.
\newblock The church-turing thesis.
\newblock In Zalta, E. (ed.), \emph{Stanford Encyclopedia of Philosophy}. Stanford Encyclopedia of Philosophy, 2012.

\bibitem[Del{\'e}tang et~al.(2023)Del{\'e}tang, Ruoss, Duquenne, Catt, Genewein, Mattern, Grau-Moya, Wenliang, Aitchison, Orseau, et~al.]{deletang2023language}
Del{\'e}tang, G., Ruoss, A., Duquenne, P.-A., Catt, E., Genewein, T., Mattern, C., Grau-Moya, J., Wenliang, L.~K., Aitchison, M., Orseau, L., et~al.
\newblock Language modeling is compression.
\newblock \emph{arXiv preprint arXiv:2309.10668}, 2023.

\bibitem[Delétang et~al.(2024)Delétang, Ruoss, Duquenne, Catt, Genewein, Mattern, Grau-Moya, Wenliang, Aitchison, Orseau, Hutter, and Veness]{delétang2024languagemodelingcompression}
Delétang, G., Ruoss, A., Duquenne, P.-A., Catt, E., Genewein, T., Mattern, C., Grau-Moya, J., Wenliang, L.~K., Aitchison, M., Orseau, L., Hutter, M., and Veness, J.
\newblock Language modeling is compression, 2024.
\newblock URL \url{https://arxiv.org/abs/2309.10668}.

\bibitem[Desai et~al.(2015)Desai, Gulwani, Hingorani, Jain, Karkare, Marron, R, and Roy]{desai2015programsynthesisusingnatural}
Desai, A., Gulwani, S., Hingorani, V., Jain, N., Karkare, A., Marron, M., R, S., and Roy, S.
\newblock Program synthesis using natural language, 2015.
\newblock URL \url{https://arxiv.org/abs/1509.00413}.

\bibitem[Dong et~al.(2024)Dong, Li, Dai, Zheng, Ma, Li, Xia, Xu, Wu, Chang, et~al.]{dong2024survey}
Dong, Q., Li, L., Dai, D., Zheng, C., Ma, J., Li, R., Xia, H., Xu, J., Wu, Z., Chang, B., et~al.
\newblock A survey on in-context learning.
\newblock In \emph{Proceedings of the 2024 conference on empirical methods in natural language processing}, pp.\  1107--1128, 2024.

\bibitem[Dosovitskiy et~al.(2021)Dosovitskiy, Beyer, Kolesnikov, Weissenborn, Zhai, Unterthiner, Dehghani, Minderer, Heigold, Gelly, Uszkoreit, and Houlsby]{dosovitskiy2021imageworth16x16words}
Dosovitskiy, A., Beyer, L., Kolesnikov, A., Weissenborn, D., Zhai, X., Unterthiner, T., Dehghani, M., Minderer, M., Heigold, G., Gelly, S., Uszkoreit, J., and Houlsby, N.
\newblock An image is worth 16x16 words: Transformers for image recognition at scale, 2021.
\newblock URL \url{https://arxiv.org/abs/2010.11929}.

\bibitem[Evanson et~al.(2023)Evanson, Lakretz, and King]{evanson2023language}
Evanson, L., Lakretz, Y., and King, J.-R.
\newblock Language acquisition: do children and language models follow similar learning stages?
\newblock \emph{arXiv preprint arXiv:2306.03586}, 2023.

\bibitem[Field(1987)]{field1987relations}
Field, D.~J.
\newblock Relations between the statistics of natural images and the response properties of cortical cells.
\newblock \emph{Journal of the Optical Society of America A}, 4\penalty0 (12):\penalty0 2379--2394, 1987.

\bibitem[Finzi et~al.(2026)Finzi, Qiu, Jiang, Izmailov, Kolter, and Wilson]{finzi2026entropy}
Finzi, M., Qiu, S., Jiang, Y., Izmailov, P., Kolter, J.~Z., and Wilson, A.~G.
\newblock From entropy to epiplexity: Rethinking information for computationally bounded intelligence.
\newblock \emph{arXiv preprint arXiv:2601.03220}, 2026.

\bibitem[Gardner(1970)]{gardner1970mathematical}
Gardner, M.
\newblock Mathematical games.
\newblock \emph{Scientific american}, 222\penalty0 (6):\penalty0 132--140, 1970.

\bibitem[Garg et~al.(2023)Garg, Tsipras, Liang, and Valiant]{garg2023transformerslearnincontextcase}
Garg, S., Tsipras, D., Liang, P., and Valiant, G.
\newblock What can transformers learn in-context? a case study of simple function classes, 2023.
\newblock URL \url{https://arxiv.org/abs/2208.01066}.

\bibitem[Geirhos et~al.(2020)Geirhos, Jacobsen, Michaelis, Zemel, Brendel, Bethge, and Wichmann]{geirhos2020shortcut}
Geirhos, R., Jacobsen, J.-H., Michaelis, C., Zemel, R., Brendel, W., Bethge, M., and Wichmann, F.~A.
\newblock Shortcut learning in deep neural networks.
\newblock \emph{Nature Machine Intelligence}, 2\penalty0 (11):\penalty0 665--673, 2020.

\bibitem[Gokaslan et~al.(2019)Gokaslan, Cohen, Pavlick, and Tellex]{Gokaslan2019OpenWeb}
Gokaslan, A., Cohen, V., Pavlick, E., and Tellex, S.
\newblock Openwebtext corpus.
\newblock \url{http://Skylion007.github.io/OpenWebTextCorpus}, 2019.

\bibitem[Han et~al.(2025{\natexlab{a}})Han, Pari, Gershman, and Agrawal]{han2025generalintelligencerequiresrewardbased}
Han, S., Pari, J., Gershman, S.~J., and Agrawal, P.
\newblock General intelligence requires reward-based pretraining, 2025{\natexlab{a}}.
\newblock URL \url{https://arxiv.org/abs/2502.19402}.

\bibitem[Han et~al.(2025{\natexlab{b}})Han, Song, Gore, and Agrawal]{han2025emergenceeffectivenesstaskvectors}
Han, S., Song, J., Gore, J., and Agrawal, P.
\newblock Emergence and effectiveness of task vectors in in-context learning: An encoder decoder perspective, 2025{\natexlab{b}}.
\newblock URL \url{https://arxiv.org/abs/2412.12276}.

\bibitem[Hoffmann et~al.(2022)Hoffmann, Borgeaud, Mensch, Buchatskaya, Cai, Rutherford, de~Las~Casas, Hendricks, Welbl, Clark, Hennigan, Noland, Millican, van~den Driessche, Damoc, Guy, Osindero, Simonyan, Elsen, Rae, Vinyals, and Sifre]{hoffmann2022trainingcomputeoptimallargelanguage}
Hoffmann, J., Borgeaud, S., Mensch, A., Buchatskaya, E., Cai, T., Rutherford, E., de~Las~Casas, D., Hendricks, L.~A., Welbl, J., Clark, A., Hennigan, T., Noland, E., Millican, K., van~den Driessche, G., Damoc, B., Guy, A., Osindero, S., Simonyan, K., Elsen, E., Rae, J.~W., Vinyals, O., and Sifre, L.
\newblock Training compute-optimal large language models, 2022.
\newblock URL \url{https://arxiv.org/abs/2203.15556}.

\bibitem[Hu et~al.(2025{\natexlab{a}})Hu, Petty, Shi, Merrill, and Linzen]{hu2025between}
Hu, M.~Y., Petty, J., Shi, C., Merrill, W., and Linzen, T.
\newblock Between circuits and chomsky: Pre-pretraining on formal languages imparts linguistic biases.
\newblock \emph{arXiv preprint arXiv:2502.19249}, 2025{\natexlab{a}}.

\bibitem[Hu et~al.(2025{\natexlab{b}})Hu, Petty, Shi, Merrill, and Linzen]{hu2025circuitschomskyprepretrainingformal}
Hu, M.~Y., Petty, J., Shi, C., Merrill, W., and Linzen, T.
\newblock Between circuits and chomsky: Pre-pretraining on formal languages imparts linguistic biases, 2025{\natexlab{b}}.
\newblock URL \url{https://arxiv.org/abs/2502.19249}.

\bibitem[Huh et~al.(2024)Huh, Cheung, Wang, and Isola]{huh2024platonicrepresentationhypothesis}
Huh, M., Cheung, B., Wang, T., and Isola, P.
\newblock The platonic representation hypothesis, 2024.
\newblock URL \url{https://arxiv.org/abs/2405.07987}.

\bibitem[Jaech et~al.(2024)Jaech, Kalai, Lerer, Richardson, El-Kishky, Low, Helyar, Madry, Beutel, Carney, et~al.]{jaech2024openai}
Jaech, A., Kalai, A., Lerer, A., Richardson, A., El-Kishky, A., Low, A., Helyar, A., Madry, A., Beutel, A., Carney, A., et~al.
\newblock Openai o1 system card.
\newblock \emph{arXiv preprint arXiv:2412.16720}, 2024.

\bibitem[Jelassi et~al.(2025)Jelassi, Mohri, Brandfonbrener, Gu, Vyas, Anand, Alvarez-Melis, Li, Kakade, and Malach]{jelassi2025mixtureparrotsexpertsimprove}
Jelassi, S., Mohri, C., Brandfonbrener, D., Gu, A., Vyas, N., Anand, N., Alvarez-Melis, D., Li, Y., Kakade, S.~M., and Malach, E.
\newblock Mixture of parrots: Experts improve memorization more than reasoning, 2025.
\newblock URL \url{https://arxiv.org/abs/2410.19034}.

\bibitem[Jiang(2023)]{jiang2023latent}
Jiang, H.
\newblock A latent space theory for emergent abilities in large language models.
\newblock \emph{arXiv preprint arXiv:2304.09960}, 2023.

\bibitem[Kaplan et~al.(2020)Kaplan, McCandlish, Henighan, Brown, Chess, Child, Gray, Radford, Wu, and Amodei]{kaplan2020scalinglawsneurallanguage}
Kaplan, J., McCandlish, S., Henighan, T., Brown, T.~B., Chess, B., Child, R., Gray, S., Radford, A., Wu, J., and Amodei, D.
\newblock Scaling laws for neural language models, 2020.
\newblock URL \url{https://arxiv.org/abs/2001.08361}.

\bibitem[Kataoka et~al.(2020)Kataoka, Okayasu, Matsumoto, Yamagata, Yamada, Inoue, Nakamura, and Satoh]{kataoka2020pre}
Kataoka, H., Okayasu, K., Matsumoto, A., Yamagata, E., Yamada, R., Inoue, N., Nakamura, A., and Satoh, Y.
\newblock Pre-training without natural images.
\newblock In \emph{Proceedings of the Asian Conference on Computer Vision}, 2020.

\bibitem[Kirsch et~al.(2022)Kirsch, Harrison, Sohl-Dickstein, and Metz]{kirsch2022general}
Kirsch, L., Harrison, J., Sohl-Dickstein, J., and Metz, L.
\newblock General-purpose in-context learning by meta-learning transformers.
\newblock \emph{arXiv preprint arXiv:2212.04458}, 2022.

\bibitem[Kolmogorov(1963)]{kolmogorov1963tables}
Kolmogorov, A.~N.
\newblock On tables of random numbers.
\newblock \emph{Sankhy{\=a}: The Indian Journal of Statistics, Series A}, pp.\  369--376, 1963.

\bibitem[Li et~al.(2024)Li, Hou, Sachan, and Cotterell]{li2024languagemodelslearncontext}
Li, J., Hou, Y., Sachan, M., and Cotterell, R.
\newblock What do language models learn in context? the structured task hypothesis, 2024.
\newblock URL \url{https://arxiv.org/abs/2406.04216}.

\bibitem[Li et~al.(2025)Li, Guo, Yang, Xu, Wu, and He]{li2025codeiocondensingreasoningpatterns}
Li, J., Guo, D., Yang, D., Xu, R., Wu, Y., and He, J.
\newblock Codei/o: Condensing reasoning patterns via code input-output prediction, 2025.
\newblock URL \url{https://arxiv.org/abs/2502.07316}.

\bibitem[Li \& Vit{\'a}nyi(2019)Li and Vit{\'a}nyi]{LiVitanyi2019KC4}
Li, M. and Vit{\'a}nyi, P.
\newblock \emph{An Introduction to Kolmogorov Complexity and Its Applications}.
\newblock Texts in Computer Science. Springer, Cham, 4 edition, 2019.
\newblock ISBN 978-3-030-11297-4.
\newblock \doi{10.1007/978-3-030-11298-1}.
\newblock URL \url{https://doi.org/10.1007/978-3-030-11298-1}.

\bibitem[Li et~al.(2023)Li, Bubeck, Eldan, Giorno, Gunasekar, and Lee]{li2023textbooksneediiphi15}
Li, Y., Bubeck, S., Eldan, R., Giorno, A.~D., Gunasekar, S., and Lee, Y.~T.
\newblock Textbooks are all you need ii: phi-1.5 technical report, 2023.
\newblock URL \url{https://arxiv.org/abs/2309.05463}.

\bibitem[Lloyd(2001)]{lloyd2001measures}
Lloyd, S.
\newblock Measures of complexity: a nonexhaustive list.
\newblock \emph{IEEE Control Systems Magazine}, 21\penalty0 (4):\penalty0 7--8, 2001.

\bibitem[Lozhkov et~al.(2024)Lozhkov, Ben~Allal, von Werra, and Wolf]{fwe}
Lozhkov, A., Ben~Allal, L., von Werra, L., and Wolf, T.
\newblock Fineweb-edu: the finest collection of educational content, 2024.
\newblock URL \url{https://huggingface.co/datasets/HuggingFaceFW/fineweb-edu}.

\bibitem[Lu et~al.(2022)Lu, Grover, Abbeel, and Mordatch]{lu2022frozen}
Lu, K., Grover, A., Abbeel, P., and Mordatch, I.
\newblock Frozen pretrained transformers as universal computation engines.
\newblock In \emph{Proceedings of the AAAI conference on artificial intelligence}, volume~36, pp.\  7628--7636, 2022.

\bibitem[Lu et~al.(2024)Lu, Zhou, Ren, Wang, Shi, Pan, Zhan, and Li]{lu2024mathgeniegeneratingsyntheticdata}
Lu, Z., Zhou, A., Ren, H., Wang, K., Shi, W., Pan, J., Zhan, M., and Li, H.
\newblock Mathgenie: Generating synthetic data with question back-translation for enhancing mathematical reasoning of llms, 2024.
\newblock URL \url{https://arxiv.org/abs/2402.16352}.

\bibitem[Mirchandani et~al.(2023)Mirchandani, Xia, Florence, Ichter, Driess, Arenas, Rao, Sadigh, and Zeng]{mirchandani2023large}
Mirchandani, S., Xia, F., Florence, P., Ichter, B., Driess, D., Arenas, M.~G., Rao, K., Sadigh, D., and Zeng, A.
\newblock Large language models as general pattern machines.
\newblock \emph{arXiv preprint arXiv:2307.04721}, 2023.

\bibitem[Mitchell(2009)]{mitchell2009complexity}
Mitchell, M.
\newblock \emph{Complexity: A guided tour}.
\newblock Oxford University Press, 2009.

\bibitem[Mordvintsev et~al.(2020)Mordvintsev, Randazzo, Niklasson, and Levin]{mordvintsev2020growing}
Mordvintsev, A., Randazzo, E., Niklasson, E., and Levin, M.
\newblock Growing neural cellular automata.
\newblock \emph{Distill}, 2020.
\newblock \doi{10.23915/distill.00023}.
\newblock https://distill.pub/2020/growing-ca.

\bibitem[Mota et~al.(2013)Mota, Aaronson, Antunes, and Souto]{mota2013sophistication}
Mota, F., Aaronson, S., Antunes, L., and Souto, A.
\newblock Sophistication as randomness deficiency.
\newblock In \emph{Descriptional Complexity of Formal Systems: 15th International Workshop, DCFS 2013, London, ON, Canada, July 22-25, 2013. Proceedings 15}, pp.\  172--181. Springer, 2013.

\bibitem[Mukherjee et~al.(2023)Mukherjee, Mitra, Jawahar, Agarwal, Palangi, and Awadallah]{mukherjee2023orcaprogressivelearningcomplex}
Mukherjee, S., Mitra, A., Jawahar, G., Agarwal, S., Palangi, H., and Awadallah, A.
\newblock Orca: Progressive learning from complex explanation traces of gpt-4, 2023.
\newblock URL \url{https://arxiv.org/abs/2306.02707}.

\bibitem[Nadǎş et~al.(2025)Nadǎş, Dioşan, and Tomescu]{Nad__2025}
Nadǎş, M., Dioşan, L., and Tomescu, A.
\newblock Synthetic data generation using large language models: Advances in text and code.
\newblock \emph{IEEE Access}, 13:\penalty0 134615–134633, 2025.
\newblock ISSN 2169-3536.
\newblock \doi{10.1109/access.2025.3589503}.
\newblock URL \url{http://dx.doi.org/10.1109/ACCESS.2025.3589503}.

\bibitem[Olsson et~al.(2022)Olsson, Elhage, Nanda, Joseph, DasSarma, Henighan, Mann, Askell, Bai, Chen, et~al.]{olsson2022context}
Olsson, C., Elhage, N., Nanda, N., Joseph, N., DasSarma, N., Henighan, T., Mann, B., Askell, A., Bai, Y., Chen, A., et~al.
\newblock In-context learning and induction heads.
\newblock \emph{arXiv preprint arXiv:2209.11895}, 2022.

\bibitem[Papadimitriou \& Jurafsky(2023)Papadimitriou and Jurafsky]{papadimitriou2023injecting}
Papadimitriou, I. and Jurafsky, D.
\newblock Injecting structural hints: Using language models to study inductive biases in language learning.
\newblock \emph{arXiv preprint arXiv:2304.13060}, 2023.

\bibitem[Paster et~al.(2023)Paster, Santos, Azerbayev, and Ba]{paster2023openwebmath}
Paster, K., Santos, M.~D., Azerbayev, Z., and Ba, J.
\newblock Openwebmath: An open dataset of high-quality mathematical web text, 2023.

\bibitem[Polyanskiy \& Wu(2025)Polyanskiy and Wu]{Polyanskiy_Wu_2025}
Polyanskiy, Y. and Wu, Y.
\newblock \emph{Information Theory: From Coding to Learning}.
\newblock Cambridge University Press, 2025.

\bibitem[Reid et~al.(2022)Reid, Yamada, and Gu]{reid2022can}
Reid, M., Yamada, Y., and Gu, S.~S.
\newblock Can wikipedia help offline reinforcement learning?
\newblock \emph{arXiv preprint arXiv:2201.12122}, 2022.

\bibitem[Rendell(2002)]{Rendell2002}
Rendell, P.
\newblock \emph{Turing Universality of the Game of Life}, pp.\  513--539.
\newblock Springer London, London, 2002.
\newblock ISBN 978-1-4471-0129-1.
\newblock \doi{10.1007/978-1-4471-0129-1_18}.
\newblock URL \url{https://doi.org/10.1007/978-1-4471-0129-1_18}.

\bibitem[Ribeiro et~al.(2023)Ribeiro, Bernardes, and Mello]{fractal-language}
Ribeiro, L.~C., Bernardes, A.~T., and Mello, H.
\newblock On the fractal patterns of language structures.
\newblock \emph{PLOS ONE}, 18\penalty0 (5):\penalty0 1--20, 05 2023.
\newblock \doi{10.1371/journal.pone.0285630}.
\newblock URL \url{https://doi.org/10.1371/journal.pone.0285630}.

\bibitem[Ruis et~al.(2024)Ruis, Mozes, Bae, Kamalakara, Talupuru, Locatelli, Kirk, Rockt{\"a}schel, Grefenstette, and Bartolo]{ruis2024procedural}
Ruis, L., Mozes, M., Bae, J., Kamalakara, S.~R., Talupuru, D., Locatelli, A., Kirk, R., Rockt{\"a}schel, T., Grefenstette, E., and Bartolo, M.
\newblock Procedural knowledge in pretraining drives reasoning in large language models.
\newblock \emph{arXiv preprint arXiv:2411.12580}, 2024.

\bibitem[Saxton et~al.(2019)Saxton, Grefenstette, Hill, and Kohli]{saxton2019analysingmathematicalreasoningabilities}
Saxton, D., Grefenstette, E., Hill, F., and Kohli, P.
\newblock Analysing mathematical reasoning abilities of neural models, 2019.
\newblock URL \url{https://arxiv.org/abs/1904.01557}.

\bibitem[Sharma et~al.(2023)Sharma, Cz{\'e}gel, Lachmann, Kempes, Walker, and Cronin]{sharma2023assembly}
Sharma, A., Cz{\'e}gel, D., Lachmann, M., Kempes, C.~P., Walker, S.~I., and Cronin, L.
\newblock Assembly theory explains and quantifies selection and evolution.
\newblock \emph{Nature}, 622\penalty0 (7982):\penalty0 321--328, 2023.

\bibitem[Shinnick et~al.(2025{\natexlab{a}})Shinnick, Jiang, Saratchandran, Hengel, and Teney]{shinnick2025transformers}
Shinnick, Z., Jiang, L., Saratchandran, H., Hengel, A. v.~d., and Teney, D.
\newblock Transformers pretrained on procedural data contain modular structures for algorithmic reasoning.
\newblock \emph{arXiv preprint arXiv:2505.22308}, 2025{\natexlab{a}}.

\bibitem[Shinnick et~al.(2025{\natexlab{b}})Shinnick, Jiang, Saratchandran, Teney, and van~den Hengel]{shinnick2025learnimagesproceduralwarmup}
Shinnick, Z., Jiang, L., Saratchandran, H., Teney, D., and van~den Hengel, A.
\newblock Can you learn to see without images? procedural warm-up for vision transformers, 2025{\natexlab{b}}.
\newblock URL \url{https://arxiv.org/abs/2511.13945}.

\bibitem[Srivastava et~al.(2023)Srivastava, Rastogi, Rao, Shoeb, Abid, Fisch, Brown, Santoro, Gupta, Garriga-Alonso, et~al.]{srivastava2023beyond}
Srivastava, A., Rastogi, A., Rao, A., Shoeb, A. A.~M., Abid, A., Fisch, A., Brown, A.~R., Santoro, A., Gupta, A., Garriga-Alonso, A., et~al.
\newblock Beyond the imitation game: Quantifying and extrapolating the capabilities of language models.
\newblock \emph{Transactions on machine learning research}, 2023.

\bibitem[Touvron et~al.(2023)Touvron, Lavril, Izacard, Martinet, Lachaux, Lacroix, Rozière, Goyal, Hambro, Azhar, Rodriguez, Joulin, Grave, and Lample]{touvron2023llamaopenefficientfoundation}
Touvron, H., Lavril, T., Izacard, G., Martinet, X., Lachaux, M.-A., Lacroix, T., Rozière, B., Goyal, N., Hambro, E., Azhar, F., Rodriguez, A., Joulin, A., Grave, E., and Lample, G.
\newblock Llama: Open and efficient foundation language models, 2023.
\newblock URL \url{https://arxiv.org/abs/2302.13971}.

\bibitem[Trockman \& Kolter(2023)Trockman and Kolter]{trockman2023mimeticinitializationselfattentionlayers}
Trockman, A. and Kolter, J.~Z.
\newblock Mimetic initialization of self-attention layers, 2023.
\newblock URL \url{https://arxiv.org/abs/2305.09828}.

\bibitem[Tunstall et~al.(2022)Tunstall, Von~Werra, and Wolf]{tunstall2022natural}
Tunstall, L., Von~Werra, L., and Wolf, T.
\newblock \emph{Natural language processing with transformers}.
\newblock " O'Reilly Media, Inc.", 2022.

\bibitem[Villalobos et~al.(2022)Villalobos, Ho, Sevilla, Besiroglu, Heim, and Hobbhahn]{villalobos2022will}
Villalobos, P., Ho, A., Sevilla, J., Besiroglu, T., Heim, L., and Hobbhahn, M.
\newblock Will we run out of data? limits of llm scaling based on human-generated data.
\newblock \emph{arXiv preprint arXiv:2211.04325}, 2022.

\bibitem[Wang et~al.(2023)Wang, Kordi, Mishra, Liu, Smith, Khashabi, and Hajishirzi]{wang2023selfinstructaligninglanguagemodels}
Wang, Y., Kordi, Y., Mishra, S., Liu, A., Smith, N.~A., Khashabi, D., and Hajishirzi, H.
\newblock Self-instruct: Aligning language models with self-generated instructions, 2023.
\newblock URL \url{https://arxiv.org/abs/2212.10560}.

\bibitem[Wei \& Zou(2019)Wei and Zou]{wei2019edaeasydataaugmentation}
Wei, J. and Zou, K.
\newblock Eda: Easy data augmentation techniques for boosting performance on text classification tasks, 2019.
\newblock URL \url{https://arxiv.org/abs/1901.11196}.

\bibitem[Wei et~al.(2022)Wei, Tay, Bommasani, Raffel, Zoph, Borgeaud, Yogatama, Bosma, Zhou, Metzler, Chi, Hashimoto, Vinyals, Liang, Dean, and Fedus]{wei2022emergentabilitieslargelanguage}
Wei, J., Tay, Y., Bommasani, R., Raffel, C., Zoph, B., Borgeaud, S., Yogatama, D., Bosma, M., Zhou, D., Metzler, D., Chi, E.~H., Hashimoto, T., Vinyals, O., Liang, P., Dean, J., and Fedus, W.
\newblock Emergent abilities of large language models, 2022.
\newblock URL \url{https://arxiv.org/abs/2206.07682}.

\bibitem[Wolfram(1984)]{WOLFRAM19841}
Wolfram, S.
\newblock Universality and complexity in cellular automata.
\newblock \emph{Physica D: Nonlinear Phenomena}, 10\penalty0 (1):\penalty0 1--35, 1984.
\newblock ISSN 0167-2789.
\newblock \doi{https://doi.org/10.1016/0167-2789(84)90245-8}.
\newblock URL \url{https://www.sciencedirect.com/science/article/pii/0167278984902458}.

\bibitem[Wolfram \& Gad-el Hak(2003)Wolfram and Gad-el Hak]{wolfram2003new}
Wolfram, S. and Gad-el Hak, M.
\newblock A new kind of science.
\newblock \emph{Appl. Mech. Rev.}, 56\penalty0 (2):\penalty0 B18--B19, 2003.

\bibitem[Wu et~al.(2025)Wu, Li, Li, Fu, Feng, Ye, Xiong, and Wang]{wu2025generatorlongcontextgenerativegenomic}
Wu, W., Li, Q., Li, M., Fu, K., Feng, F., Ye, J., Xiong, H., and Wang, Z.
\newblock Generator: A long-context generative genomic foundation model, 2025.
\newblock URL \url{https://arxiv.org/abs/2502.07272}.

\bibitem[Xie et~al.(2022)Xie, Raghunathan, Liang, and Ma]{xie2022explanationincontextlearningimplicit}
Xie, S.~M., Raghunathan, A., Liang, P., and Ma, T.
\newblock An explanation of in-context learning as implicit bayesian inference, 2022.
\newblock URL \url{https://arxiv.org/abs/2111.02080}.

\bibitem[Xu et~al.(2025)Xu, Sun, Zheng, Geng, Zhao, Feng, Tao, Lin, and Jiang]{xu2025wizardlmempoweringlargepretrained}
Xu, C., Sun, Q., Zheng, K., Geng, X., Zhao, P., Feng, J., Tao, C., Lin, Q., and Jiang, D.
\newblock Wizardlm: Empowering large pre-trained language models to follow complex instructions, 2025.
\newblock URL \url{https://arxiv.org/abs/2304.12244}.

\bibitem[Zhang et~al.(2024)Zhang, Patel, Rizvi, Liu, He, Karbasi, Zappala, and van Dijk]{zhang2024intelligence}
Zhang, S., Patel, A., Rizvi, S.~A., Liu, N., He, S., Karbasi, A., Zappala, E., and van Dijk, D.
\newblock Intelligence at the edge of chaos.
\newblock \emph{arXiv preprint arXiv:2410.02536}, 2024.

\bibitem[Zhao et~al.(2024)Zhao, Saphra, and Kakade]{zhao2024distributional}
Zhao, R., Saphra, N., and Kakade, S.~M.
\newblock Distributional scaling laws for emergent capabilities.
\newblock In \emph{NeurIPS 2024 Workshop on Scientific Methods for Understanding Deep Learning}, 2024.

\bibitem[Zipf(1949)]{zipf1949human}
Zipf, G.~K.
\newblock \emph{Human behavior and the principle of least effort.}
\newblock Addison-Wesley Press, 1949.

\bibitem[Ziv \& Lempel(1977)Ziv and Lempel]{lempelziv}
Ziv, J. and Lempel, A.
\newblock A universal algorithm for sequential data compression.
\newblock \emph{IEEE Transactions on Information Theory}, 23\penalty0 (3):\penalty0 337--343, 1977.
\newblock \doi{10.1109/TIT.1977.1055714}.

\end{thebibliography}
\bibliographystyle{neurips_2026}

\newpage
\appendix
\onecolumn

\section{NCA Rule Generation and Tokenization Pseudo-code}
\label{appx:psuedocode}

We describe how the NCA rules are sampled according gzip compression ratio and are tokenized for training in the following pseudo-code in \cref{alg:nca_gen}.

\begin{algorithm}[h]
\caption{NCA Rule Generation and Tokenization}
\label{alg:nca_gen}
\begin{algorithmic}[1]
\Require Grid size $H \times W$, alphabet size $n$, patch size $p$, trajectory length $T$, temperature $\tau$, complexity threshold $r_{\min}$
\Ensure Rule parameters $\theta$, trajectory $\mathcal{X}$, token sequence $S$
\Statex
\Function{SampleComplexNCA}{$H, W, n, p, T, \tau, r_{\min}$}
    \Repeat
        \State $\theta \gets \textsc{InitNetwork}()$ \Comment{random init, frozen for the rollout}
        \State $c^{(0)} \sim \mathcal{U}\{0, \dots, n-1\}^{H \times W}$ \Comment{random initial grid}
        \For{$t = 1$ \textbf{to} $T$}
            \For{each cell $i$}
                \State $c^{(t)}_i \sim \mathrm{softmax}\!\left(f_\theta(N(c^{(t-1)}, i)) / \tau\right)$
            \EndFor
        \EndFor
        \State $\mathcal{C} \gets (c^{(0)}, c^{(1)}, \dots, c^{(T)})$
        \State $S \gets \Call{Tokenize}{\mathcal{C}, n, p}$
        \State $r \gets |\text{gzip}(S)| \,/\, |S|$ \Comment{compression ratio in $(0, 1]$}
    \Until{$r > r_{\min}$}
    \State \Return $(\theta, \mathcal{C}, S)$
\EndFunction
\Statex
\Function{Tokenize}{$\mathcal{C}, n, p$}
    \State $S \gets$ empty sequence
    \ForAll{grids $c^{(t)} \in \mathcal{C}$}
        \State Partition $c^{(t)}$ into non-overlapping $p \times p$ patches $\{p_{t,1}, \dots, p_{t,M}\}$ in row-major order, where $M = HW / p^2$
        \ForAll{patches $p_{t,j}$}
            \State $\text{token} \gets \sum_{\ell=0}^{p^2 - 1} p_{t,j}[\ell] \cdot n^{\ell}$ \Comment{base-$n$ numeral over $p^2$ cells}
            \State Append $\text{token}$ to $S$
        \EndFor
    \EndFor
    \State \Return $S$
\EndFunction
\end{algorithmic}
\end{algorithm}

\section{Analysis on Natural and Synthetic Data Distributions}
\label{app:data_distribution_analysis}

\begin{figure}[h]
    \centering
    \includegraphics[width=0.5\linewidth]{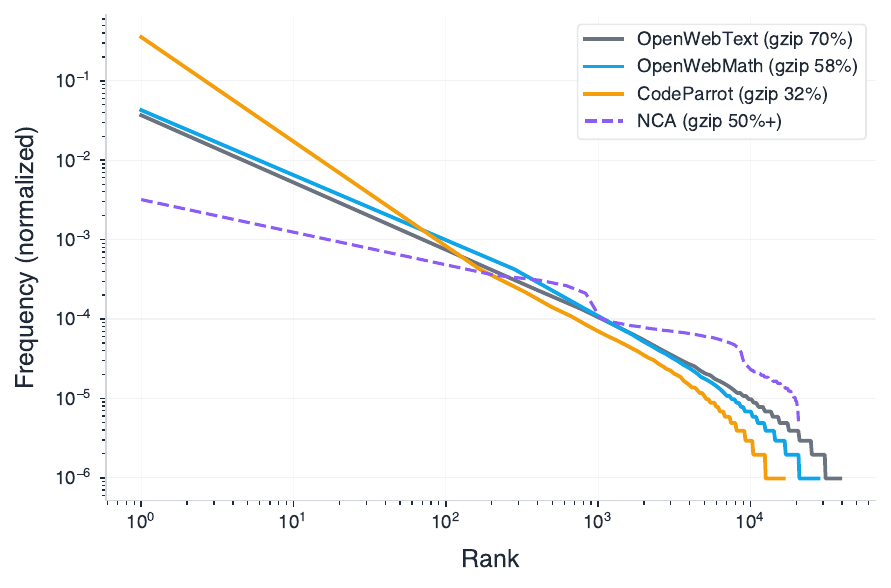}
    \caption{\textbf{NCA data exhibits a similar Zipfian or power-law structure to natural language.} We compare the relative token frequency distribution for each of the natural language corpora and NCA data. Natural language from different domains has different average complexity as measured by gzip compressibility (see legend).}
    \label{fig:token-distributions}
\end{figure}

\begin{figure}[t]
    \centering
    \includegraphics[width=\linewidth]{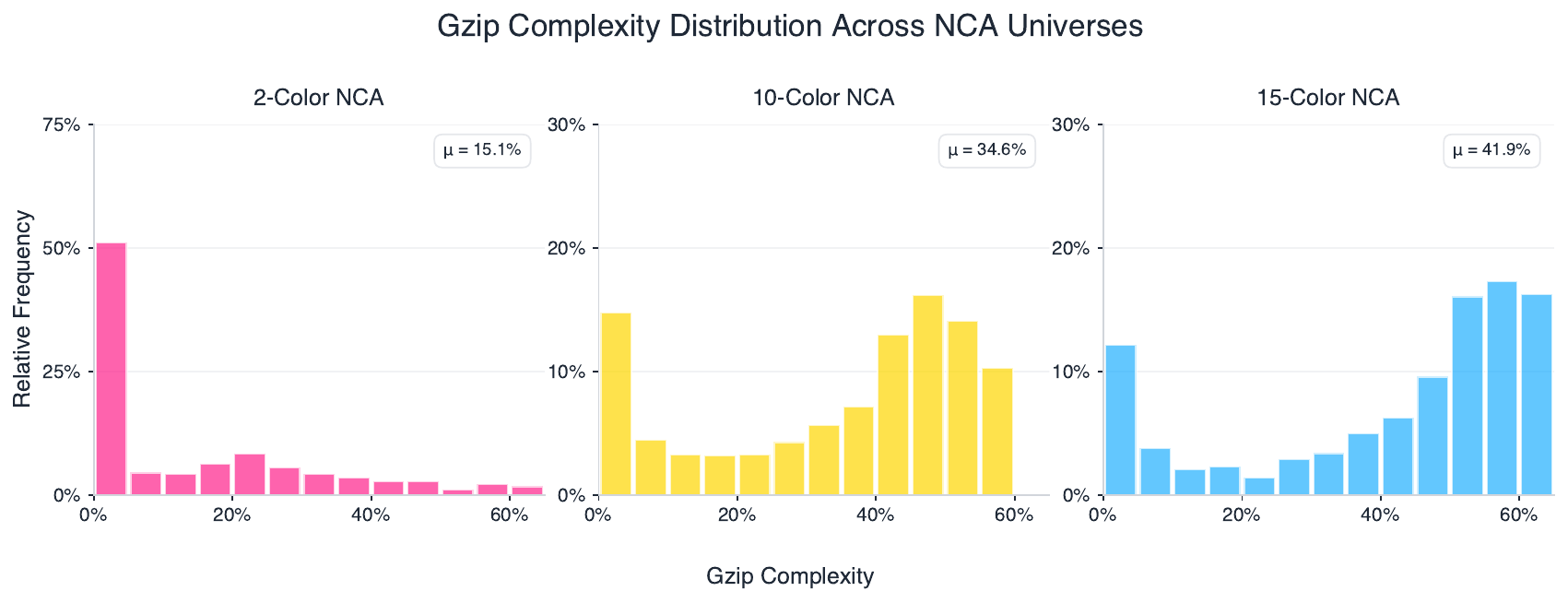}
    \caption{Different NCA alphabet sizes, $n=2,10,15$ naturally yield different complexity distributions of the data. Increasing $n$ inherently increases the complexity of the data.}
    \label{fig:nca_gzip}
\end{figure}

In this section, we examine the distributions of natural-language and NCA-generated synthetic data with respect to two primary high-order heuristics: (1) token frequency distribution and (2) gzip compressibility.
\subsection{NCA data exhibits similar token distributions to natural language}
To compare the token frequency across different data distributions, we sample and tokenize random text sequences from the natural language datasets (OpenWebText, OpenWebMath, and CodeParrot) and NCA generated data ($n=15$). Figure~\ref{fig:token-distributions} shows the distribution of relative token frequencies. Data generated from the NCAs follows a heavy-tailed, Zipfian token distribution that is structured similarly to natural language. Another interesting observation is that natural language depending on the domain varies quite drastically from 32\% in code and 60-70\% in math and web text.

\subsection{Increasing the alphabet size $n$ leads to more complex generated trajectories}
Figure~\ref{fig:nca_gzip} compares the distribution of gzip complexity across trajectories generated by randomly sampled NCAs across different alphabet sizes $n=2,10,15$. As $n$ increases, the distribution skews towards less compressible, more complex data. This implies that with higher $n$, the universe of rules expands and naturally the dynamics become more complex.

\section{Detailed Pre-pre-training and Pre-training Setup}
\label{app:experimental-setup}

For our 1.6B parameter Llama-based transformer, we use 24 layers, 32 heads, and hidden dimension of 2048. For the 3B model, we use 26 layers, 32 heads, hidden dimension of 3072.

Table~\ref{tab:hyperparams} summarizes the hyperparameters used for both pre-pre-training on NCA data and subsequent pretraining on natural language datasets. We sweep various batch sizes (32 to 512), learning rates ($1\times10^{-3}$ to $1\times10^{-5}$), and weight decays ($1\times10^{-1}$ to $1\times 10^{-6}$). To ensure reproducibility, we train our pipeline on 4 randomness seeds for each main pipeline (NCA Pre-pre-training, Scratch, and C4 Pre-pre-training) and at least 2 seeds for each ablation run.

We train all models on several nodes of H200, encompassing~20,000+ compute hours across main results and ablations. NCA pre-pretraining data generation cost is negligible, as the generator model is parameterized by a small network that can be run on a CPU to generate the full dataset within minutes.

\begin{table}[h]
\centering
\small
\renewcommand{\arraystretch}{1.15}
\resizebox{\columnwidth}{!}{%
\begin{tabular}{lccc}
\toprule
\textbf{Hyperparameter} & \textbf{Pre-pre-training} & \multicolumn{2}{c}{\textbf{Pre-training}} \\
\cmidrule(lr){3-4}
& & \textbf{1.6B} & \textbf{3B} \\
\midrule
Effective batch size & 32 & 512 & 2048 \\
Sequence length & 1024 (1.6B) or 2048 (3B) tokens & 1024 tokens & 2048 tokens \\
Learning rate & $1 \times 10^{-4}$ & $5 \times 10^{-4}$ (Math / OpenWebText / Fineweb-Edu), & $1\times 10^{-4}$ (Fineweb-Edu-3B) \\
& & $2\times 10^{-4}$ (Code) & $3\times10^{-4}$ (OpenWebText) \\
LR schedule & Cosine w/ warmup & Cosine w/ warmup & Cosine w/ warmup \\
Warmup steps (\% total) & 10\% & 10\% & 10\% \\
Optimizer & AdamW & AdamW & AdamW \\
Weight decay & None & $1 \times 10^{-4}$ & $1\times 10^{-1}$ \\
Gradient clipping & None & 1.0 & 1.0 \\
\bottomrule
\end{tabular}
}
\caption{Hyperparameters for pre-pre-training and pre-training experiments.}
\label{tab:hyperparams}
\end{table}
\section{Detailed Fine-Tuning Setup}
\label{app:instruction-ft-setup}
For GSM8K and HumanEval, we evaluate on all tasks provided by the benchmarks. For BigBench-Lite, given the quantity and imbalance of samples and tasks, only tasks with above 100 samples were chosen. The exact BBL task list was: bbq\_lite\_json, formal\_fallacies\_syllogisms\_negation, logic\_grid\_puzzle, logical\_deduction, operators, play\_dialog\_same\_or\_different, strange\_stories, strategyqa, symbol\_interpretation, vitaminc\_fact\_verification, and winowhy

For GSM8K and Big Bench-Lite, we fine-tune the OpenWebMath and OpenWebText pre-trained models on the respective training sets for 10 epochs using a LR of 1e-5 to enable the models to follow the question answering format for evaluation. For GSM8k, we also fine-tune on the Chain-of-Thought reasoning trace provided by the dataset. FFor HumanEval, we do not fine-tune the models since it is a code completion task. For reproducibility, we train 4 seeds for each model and baseline and report the averages across runs.

\begin{wrapfigure}{r}{0.4\columnwidth}
\vskip -0.2 in
\centerline{\includegraphics[width=0.4\columnwidth]{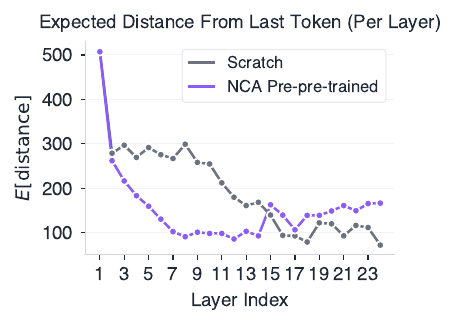}}
\vskip -0.1 in
\caption{\textbf{The expected attention distance from the last token, per layer.} The two models exhibit inverted patterns: the scratch model attends broadly in early layers and narrows with depth, whereas the NCA pre-pre-trained model contracts attention quickly before expanding it again in deeper layers.}
\label{fig:expected-attention-distance}
\vskip -0.2 in
\end{wrapfigure}

We evaluate the models' performances across different Pass@$k$. For GSM8K and Humaneval, we vary $k$ from 1, 8, 16, and 32. For BigBench, given some tasks are multiple choice, we only report $k$ from 1, 2 and 4, with normalization against random guessing. We use the unbiased estimator from \citet{chen2021evaluating}, computing the metric from 64 total decodings per run. For evaluation, we sampled HumanEval with a temperature of $0.4$ and top-p of $0.95$, GSM8K with a temperature of $0.2$ and top-p of $0.95$ and BigBench-Lite with a temperature of $0.4$ and top-p of $0.95$. We use 4 training pipeline seeds per task for each baseline and NCA pre-pretrained models and 5 decoding seeds per pipeline seed. For Pass@k evaluation, we sweep generation settings across temperatures of 0.0, 0.2, 0.4, 0.6, and 1.0, and select the best performing single temperature for each task when computing accuracies.

\section{Perplexity and Reasoning Detailed Results and Statistical Significance}
\label{app:statistical_significance}
In Table~\ref{table:combined-results} we share more detailed tables outlining the exact values and standard deviations for final validation perplexity during pretraining and accuracy at varying pass@k for reasoning benchmarks.

\begin{table*}[ht]
\centering
\scriptsize
\setlength{\tabcolsep}{3pt}
\renewcommand{\arraystretch}{1.1}

\begin{minipage}[t]{0.48\textwidth}
\centering
\vspace{0pt}
\textbf{a) Perplexity} \\
\begin{tabular*}{\linewidth}{l @{\extracolsep{\fill}} c c c c c}
\toprule
\textbf{Dataset} & \textbf{Scratch} & \textbf{NCA} & \textbf{Dyck} & \textbf{C4} & \textbf{FW-Edu} \\
\midrule
\textit{OpenWebText} & 14.66 & \textbf{13.82} & 14.36 & 14.69 & 14.66 \\
\tiny{Std Dev} & $\pm$0.02 & \textbf{$\pm$0.01} & $\pm$0.10 & $\pm$0.04 & $\pm$0.02 \\
\midrule
\textit{CodeParrot} & 1.92 & \textbf{1.84} & 1.85 & 1.89 & 1.89 \\
\tiny{Std Dev} & $\pm$0.04 & \textbf{$\pm$0.00} & $\pm$0.01 & $\pm$0.00 & $\pm$0.02 \\
\midrule
\textit{OpenWebMath} & 8.11 & \textbf{7.70} & 7.91 & 8.14 & 8.16 \\
\tiny{Std Dev} & $\pm$0.03 & \textbf{$\pm$0.02} & $\pm$0.07 & $\pm$0.01 & $\pm$0.00 \\
\bottomrule
\end{tabular*}
\end{minipage}
\hfill
\begin{minipage}[t]{0.48\textwidth}
\centering
\vspace{0pt}
\textbf{b) Reasoning Accuracy (pass@k)} \\
\begin{tabular*}{\linewidth}{l @{\extracolsep{\fill}} c c c c c}
\toprule
\textbf{pass@$k$} & \textbf{Scratch} & \textbf{NCA} & \textbf{Dyck} & \textbf{C4} & \textbf{FW-Edu} \\
\midrule
\textit{GSM8K @1} & 3.8\% & \textbf{4.4\%} & 4.1\% & 3.8\% & 3.4\% \\
\tiny{Std Dev} & $\pm$0.1\% & \textbf{$\pm$0.3\%} & $\pm$0.4\% & $\pm$0.2\% & $\pm$0.1\% \\
\textit{GSM8K @8} & 17.9\% & \textbf{19.2\%} & 18.6\% & 17.8\% & 16.8\% \\
\tiny{Std Dev} & $\pm$0.3\% & \textbf{$\pm$0.3\%} & $\pm$0.9\% & $\pm$0.5\% & $\pm$0.4\% \\
\textit{GSM8K @32} & 36.6\% & \textbf{37.9\%} & 37.4\% & 36.2\% & 35.5\% \\
\tiny{Std Dev} & $\pm$0.6\% & \textbf{$\pm$0.3\%} & $\pm$0.7\% & $\pm$0.9\% & $\pm$0.4\% \\
\midrule
\textit{HEval @1} & 6.8\% & 7.5\% & 6.9\% & 6.3\% & 6.4\% \\
\tiny{Std Dev} & $\pm$0.6\% & $\pm$0.4\% & $\pm$0.1\% & $\pm$0.3\% & $\pm$0.2\% \\
\textit{HEval @8} & 11.2\% & 11.4\% & 11.3\% & 10.5\% & 10.3\% \\
\tiny{Std Dev} & $\pm$0.6\% & $\pm$0.8\% & $\pm$0.2\% & $\pm$0.5\% & $\pm$0.2\% \\
\textit{HEval @32} & 13.9\% & 13.8\% & \textbf{14.3\%} & 12.6\% & 12.8\% \\
\tiny{Std Dev} & $\pm$0.5\% & $\pm$1.0\% & \textbf{$\pm$0.4\%} & $\pm$0.7\% & $\pm$0.2\% \\
\midrule
\textit{BB-Lite @1} & 15.4\% & 15.0\% & 13.4\% & 15.9\% & 12.7\% \\
\tiny{Std Dev} & $\pm$1.1\% & $\pm$1.2\% & $\pm$2.8\% & $\pm$0.9\% & $\pm$1.5\% \\
\textit{BB-Lite @2} & 20.9\% & \textbf{26.5\%} & 18.1\% & 22.8\% & 23.8\% \\
\tiny{Std Dev} & $\pm$2.5\% & \textbf{$\pm$1.0\%} & $\pm$2.3\% & $\pm$1.2\% & $\pm$1.6\% \\
\textit{BB-Lite @4} & 25.9\% & \textbf{36.5\%} & 22.7\% & 29.7\% & 33.5\% \\
\tiny{Std Dev} & $\pm$3.9\% & \textbf{$\pm$2.1\%} & $\pm$2.1\% & $\pm$1.3\% & $\pm$1.7\% \\

\bottomrule
\end{tabular*}
\end{minipage}

\vspace{6pt}
\caption{\textbf{NCA pre-pre-training outperforms natural and synthetic baselines across language modeling with modest downstream reasoning gains.} a) Final validation perplexity (lower is better). b) Downstream reasoning performance (higher is better). All baselines use 4 seeds of pre-pretraining and pre-training. \textbf{Bold} indicates results statistically significant ($p < 0.05$) compared to the Scratch baseline via a one-tailed Welch's t-test ($n=4$).}
\label{table:combined-results}
\end{table*}

We also compute the statistical significance of reasoning performance of NCA pre-pre-trained language models against natural language and synthetic baselines via Welch's one-tailed T-Test for $n=4$. As seen in Table~\ref{table:statistical-significance}, though gains across all reasoning benchmarks are modest, NCA pre-pre-training performance on GSM8K is statistically significant compared to all natural language baselines and performance on BigBench is statistically significant compared to all baselines at pass@2 and pass@4. This supports that NCA pre-pre-training can have a modestly positive impact on downstream reasoning performance.

\begin{table*}[ht]
\centering
\scriptsize
\setlength{\tabcolsep}{3pt}
\renewcommand{\arraystretch}{1.2}
\begin{minipage}[t]{0.48\textwidth}
\centering
\vspace{0pt}
\textbf{a) Perplexity $p$-values of NCA vs.} \\
\begin{tabular*}{\linewidth}{l @{\extracolsep{\fill}} c c c c}
\toprule
\textbf{Dataset} & \textbf{vs. Scratch} & \textbf{vs. Dyck} & \textbf{vs. C4} & \textbf{vs. FWE} \\
\midrule
\textit{OpenWebText} & \textbf{< 0.0001} & \textbf{0.0008} & \textbf{< 0.0001} & \textbf{< 0.0001} \\
\textit{CodeParrot}  & \textbf{0.0132} & \textbf{0.0159} & \textbf{< 0.0001} & \textbf{< 0.0001} \\
\textit{OpenWebMath} & \textbf{< 0.0001} & \textbf{< 0.0001} & \textbf{< 0.0001} & \textbf{< 0.0001} \\
\bottomrule
\end{tabular*}
\end{minipage}
\hfill
\begin{minipage}[t]{0.50\textwidth}
\centering
\vspace{0pt}
\textbf{b) Reasoning Accuracy $p$-values of NCA vs.} \\
\begin{tabular*}{\linewidth}{l @{\extracolsep{\fill}} c c c c}
\toprule
\textbf{Task / @k} & \textbf{vs. Scratch} & \textbf{vs. Dyck} & \textbf{vs. C4} & \textbf{vs. FWE} \\
\midrule
\textit{GSM8K @1}   & \textbf{0.0113} & 0.1394 & \textbf{0.0097} & \textbf{0.0084} \\
\textit{GSM8K @8}   & \textbf{0.0004} & 0.1402 & \textbf{0.0026} & \textbf{0.0011} \\
\textit{GSM8K @32}  & \textbf{0.0074} & 0.1292 & \textbf{0.0134} & \textbf{0.0052} \\
\midrule
\textit{HEval @1}   & 0.0537 & \textbf{0.0269} & \textbf{0.0018} & \textbf{0.0122} \\
\textit{HEval @8}   & 0.3520 & 0.4112 & 0.0571 & 0.0841 \\
\textit{HEval @32}  & N/A & N/A & \textbf{0.0115} & \textbf{0.0314} \\
\midrule
\textit{BB-Lite @1} & N/A & \textbf{0.0241} & N/A & \textbf{0.0267} \\
\textit{BB-Lite @2} & \textbf{0.0073} & \textbf{0.0012} & \textbf{0.0017} & \textbf{0.0186} \\
\textit{BB-Lite @4} & \textbf{0.0031} & \textbf{< 0.0001} & \textbf{0.0013} & \textbf{0.0365} \\
\bottomrule
\end{tabular*}
\end{minipage}
\caption{\textbf{Significance Tests (One-tailed Welch's t-test, n=4) comparing NCA Pre-pre-training against each baseline.} a) $p$-values for final validation perplexity improvements. b) $p$-values for downstream reasoning gains. \textbf{Bold} indicates statistically significant results ($p < 0.05$). N/A indicates the comparison baseline outperformed NCA for that specific metric.}
\label{table:statistical-significance}
\end{table*}

\section{Attention Comparison Between Scratch and NCA Pre-Pretrained Models}
\label{app:attention_analysis}

\begin{figure}[h]
    \centering
    \includegraphics[width=0.8\linewidth]{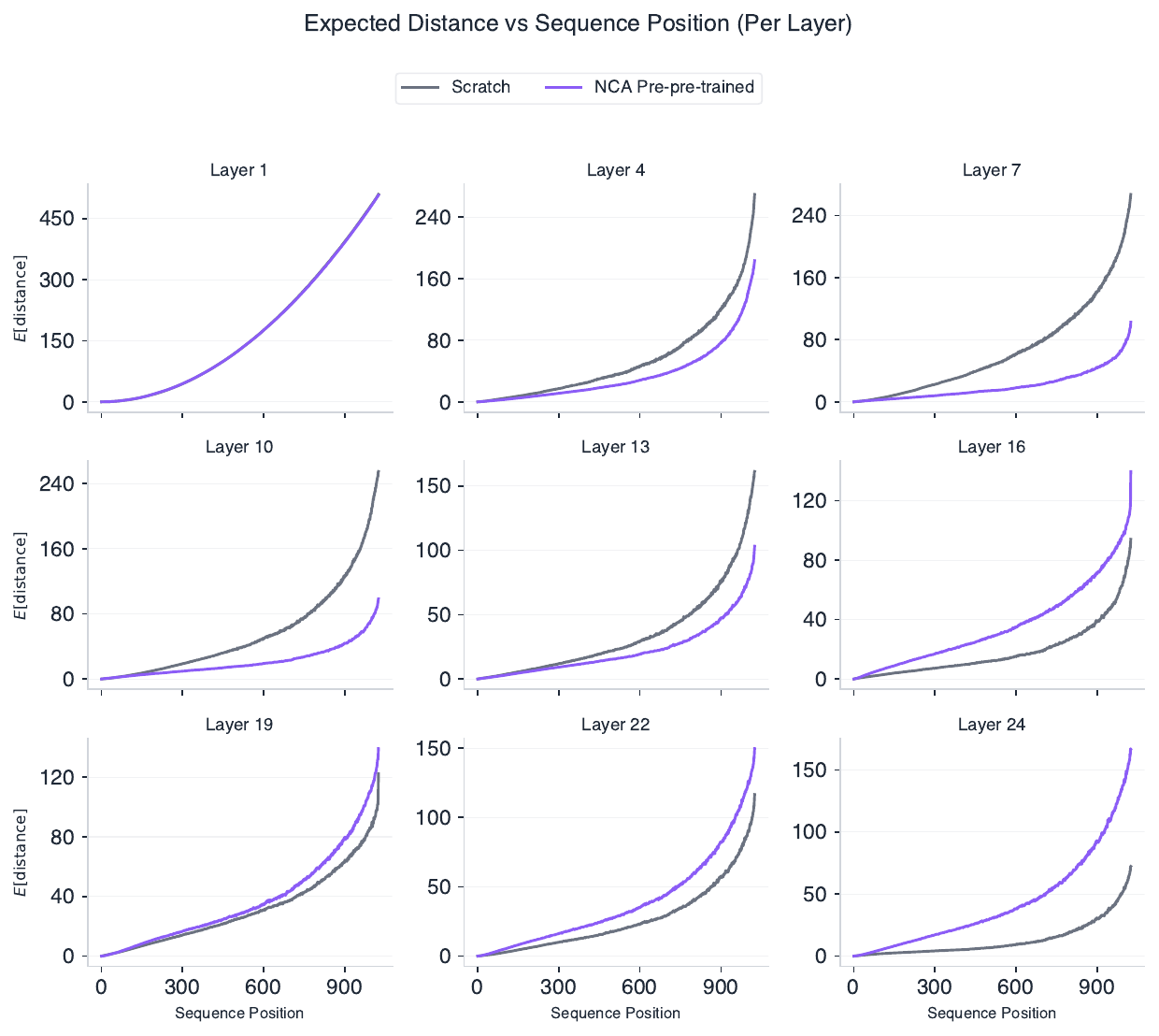}
    \caption{Expected attention distance vs. sequence position, per layer. Each panel shows $\mathbb{E}d = \sum_{j \leq i} (i - j) \cdot \alpha_{ij}$, the average distance from query position $i$
 to its attended keys, computed by averaging attention weights across heads. Curves are shown for a model trained from scratch (gray) and one initialized from NCA pre-pre-training (violet). The NCA-initialized model exhibits markedly more local attention patterns across earlier layers and more global patterns in alter layers, suggesting that pre-pre-training induces a different structural bias toward short-range and global dependencies that persists through downstream training. }
    \label{fig:attn_dist_by_layer}
\end{figure}

We provide more detailed analyses of attention patterns across scratch and NCA Pre-Pretrained models in this section.

To understand how NCA pre-pre-training shapes attention behavior, we measure the \emph{expected attention distance} from the last token at each layer: $\mathbb{E}[\text{distance}] = \sum_{d=0}^{T} d \cdot \mathrm{softmax}(\mathbf{q}^\intercal \mathbf{k}_{t-d})$,
where $\mathbf{q}$ is the query vector at position $t$, $\mathbf{k}_{t-d}$ is the key vector at position $t-d$, and $T$ is the context length. Figure~\ref{fig:expected-attention-distance} reveals a striking inversion: the scratch model attends globally in the earlier layers and locally in the later layers (15+), while the NCA pre-pre-trained model does the opposite---local first, global later.

 In Figure \ref{fig:attn_dist_by_layer}, we visualized the expected distance only from the last token, and we extend this analysis to compare the expected distance across the whole sequence length and layers. From these visualizations, we further corroborate that the NCA pre-pre-trained and scratch attention layers indeed learn fundamentally different behaviors and local/global biases of where to attend in the earlier and latter layers of the model.

\section{Additional Coding Benchmark: Mostly Basic Python Programs (MBPP)}
\label{app:coding_benchmarks}
In this section we provide details on an additional coding benchmark, Mostly Basic Python Programs, used to evaluate Codeparrot pretrained models of 1.6B parameters. As seen in Table~\ref{tab:mbpp-results}, NCA models significantly outperform scratch models across various pass@k on the benchmark, albeit absolute performance is low across both models.

\begin{table}[tbp]
\centering
\small
\caption{Pass@$k$ performance on MBPP. Results report the mean and standard
deviation across runs. $\Delta$ denotes the absolute percentage-point
improvement of NCA over training from scratch.}
\label{tab:mbpp-results}

\begin{tabular*}{\linewidth}{
    l @{\extracolsep{\fill}} c c c c
}
\toprule
\textbf{Metric}
& \textbf{Temperature}
& \textbf{Scratch}
& \textbf{NCA}
& \textbf{$\Delta$} \\
\midrule

\textit{MBPP @1}
& Greedy
& 2.45\%
& \textbf{4.30\%}
& \textbf{+1.85} \\
{\tiny Std. Dev.}
& --
& $\pm 0.53\%$
& \textbf{$\pm 0.48\%$}
& -- \\

\addlinespace[2pt]

\textit{MBPP @1}
& 0.8
& 1.53\%
& \textbf{2.15\%}
& \textbf{+0.62} \\
{\tiny Std. Dev.}
& --
& $\pm 0.10\%$
& \textbf{$\pm 0.55\%$}
& -- \\

\textit{MBPP @8}
& 0.8
& 7.46\%
& \textbf{10.52\%}
& \textbf{+3.05} \\
{\tiny Std. Dev.}
& --
& $\pm 0.26\%$
& \textbf{$\pm 2.42\%$}
& -- \\

\textit{MBPP @16}
& 0.8
& 11.51\%
& \textbf{15.57\%}
& \textbf{+4.06} \\
{\tiny Std. Dev.}
& --
& $\pm 0.57\%$
& \textbf{$\pm 3.21\%$}
& -- \\

\textit{MBPP @32}
& 0.8
& 16.95\%
& \textbf{21.50\%}
& \textbf{+4.55} \\
{\tiny Std. Dev.}
& --
& $\pm 1.33\%$
& \textbf{$\pm 3.71\%$}
& -- \\

\bottomrule
\end{tabular*}
\end{table}
\section{Details on Mechanistic Probe of Induction Heads on NCA Pre-Pretrained Models}
\label{app:induction_heads}

In this section, we outline the details of the mechanistic probe analysis conducted on NCA pre-pre-pretrained models. We run this analysis on four independently trained 1.6B NCA models and four architecture-matched random initializations across repeat length $L\in{16,32,48,64,96}$.

For each model and length $L$, we independently sampled $N=256$ random sequences $x^{(n)}=(x^{(n)}_1,\ldots,x^{(n)}_L)$ from the NCA-trained token vocabulary and constructed the repeated input $r^{(n)}_L=[x^{(n)};,x^{(n)}]$ together with a control $c^{(n)}_L=[x^{(n)};,y^{(n)}]$, where $y^{(n)}$ is an independently sampled non-repeating sequence. The two inputs share length and positional structure and differ only in whether the sequence repeats. For a query position $q$ in the second half of $r^{(n)}_L$, the token at $q$ previously occurred at $q-L$, so the standard +1 induction target is $k^\star(q,L)=q-L+1$: in the pattern $[A][B]\ldots[A]$, a prefix-matching head attends from the final $[A]$ to the earlier $[B]$.

For each head, we score and determine above-null induction heads as follows. Let $[\mathbf A^{(m,\ell,h)}(u)]_{q,k}$ denote the post-softmax attention probability from query $q$ to key $k$ for head $h$ in layer $\ell$ of model $m$ on input $u$. We computed the content-controlled prefix-matching score

$$s_{m,\ell,h}(L)=\frac{1}{N,\vert{}\mathcal{Q}L\vert{}}\sum{n=1}^{N}\sum_{q\in\mathcal{Q}_L}\Big(\big[\mathbf A^{(m,\ell,h)}(r^{(n)}L)\big]{q,k^\star(q,L)}-\big[\mathbf A^{(m,\ell,h)}(c^{(n)}L)\big]{q,k^\star(q,L)}\Big)$$

where $\mathcal Q_L$ is the set of eligible query positions in the second half; subtracting the control isolates repetition-contingent attention from positional or content-independent biases. At each $L$, the pooled scores of all heads from the four random-initialized models define an empirical null, and a head is counted as above-null if its score exceeds the null's 99th percentile.



\begin{figure}[h!]
    \centering
    \includegraphics[width=\linewidth]{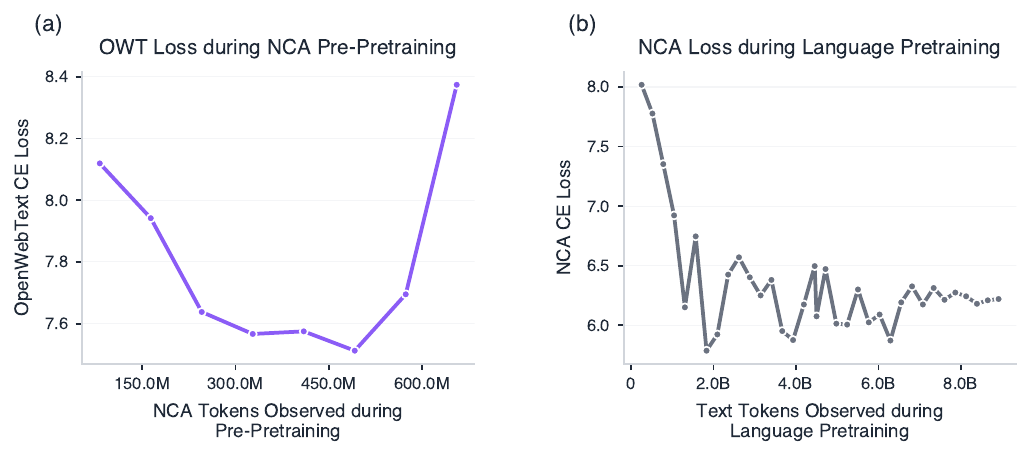}
    \vskip -0.2 in
    \caption{\textbf{NCA and OpenWebText are aligned during early stages of pre-pretraining and pre-training, but diverge later on.} \textbf{(a)} Cross-Entropy loss of model during NCA pre-pretraining on OpenWebText. \textbf{(b)} Cross-Entropy loss of scratch model during Openwebtext pretraining on NCA data. As seen in both plots, training on NCA and Openwebtext have mild initial positive impact on next token prediction on the other domain.}
    \label{fig:loss_during_train}
    \vskip -0.2 in
\end{figure}

\section{Alignment Between NCA and Natural Language}
\label{app:owt_during_nca}

In Figure~\ref{fig:loss_during_train}, we investigate the alignment of NCA-generated synthetic data with OpenWebText by analyzing how a model trained on only NCA performs on natural language (\cref{fig:loss_during_train}a), and vice versa (\cref{fig:loss_during_train}b). In both experiments, we only train on one domain (either NCA or language) and do not train on any data or import any weights from the evaluation domain. Since the size of the text tokenizer vocabulary is larger than the NCA vocabulary, for every text data sequence, we map the unique token ids, at most $2048$ (max sequence length), to the NCA vocabulary. We find surprising alignment between the two domains; Training on one domain only can lead to improvement in loss on the other domain. Furthermore, the scaling curve of loss improvement mirrors the curve we observe in Figure~\ref{fig:ablations_universe} for 10-colors. This provides further explanation via alignment for why NCA pre-pretraining may improve language modeling convergence speed and final validation perplexity.

\section{Comparison to Learning-Free Initializations}
\label{app:mimetic}

In this section, we provide details behind an additional baseline that demonstrates the benefit from NCA pre-pre-training is not just attributable to the statistics underlying the initialization of the pre-training weights. We compare the OpenWebText validation loss of a 400M model initialized using mimetic initialization \citep{trockman2023mimeticinitializationselfattentionlayers}. Mimetic initialization is a closed form learning-free initialization that biases model weights to generate structural attention patterns observed in pretrained Transformers with the aim of improving optimization and downstream performance. Concretely, it involves decomposing the self-attention query-key and value-output projection vectors as follows: $W_{Q}W_K^\intercal\approx\alpha_1 Z_1+\beta_1 I$, $W_{V}W_{\text{proj}}^\intercal\approx\alpha_2 Z_2-\beta_2 I$, where $Z_1,Z_2$ are random normal matrices and $\alpha_{1},\alpha_2,\beta_{1},\beta_2$ represent hyperparameters for the initialization. As seen in the decompositions, the resulting attention patterns have prominent positive and negative diagonals respectively.

We sweep across various configurations and values for $\alpha_1,\alpha_2,\beta_1,\beta_2\in\{0, 0.2, 0.5\}$ and report the best performing configuration. The training hyperparameters swept are the same as for the main pre-training sweep. As seen in Table~\ref{tab:mimetic}, most of the benefit of full NCA synthetic pre-pretraining cannot be recovered from this learning-free initialization, the improvement is marginal at best. We also attempt to try replicating gains by initializing the models with just first-order statistics from NCA pre-pre-trained models and also find it insufficient to cover most of the gains.

\begin{table}[t]
\centering
\caption{\textbf{Benefits of NCA Pre-pretraining cannot be recovered from learning-free mimetic initialization. }We report OpenWebText validation perplexity for 400M-parameter models.}
\label{tab:reinitialization}
\begin{tabular}{l c}
\toprule
\textbf{Method} & \textbf{Validation Perplexity $\downarrow$} \\
\midrule
Scratch & 17.83 \\
\textbf{NCA} & \textbf{16.32} \\
NCA First-Order Statistics-Based Reinitialization & 17.77 \\
Mimetic Initialization~\cite{trockman2023mimeticinitializationselfattentionlayers} & 17.81 \\
\bottomrule
\end{tabular}
\label{tab:mimetic}
\end{table}


\end{document}